\def\paperID{3506}
\def\confName{ICCV}
\def\confYear{2025}

\def\paperTitle{
CARP: Visuomotor Policy Learning \\ 
via Coarse-to-Fine Autoregressive Prediction
}

\def\authorBlock{
    Zhefei Gong$^{1}$ \qquad
    Pengxiang Ding$^{12\dagger}$ \qquad
    Shangke Lyu$^{1}$ \qquad
    Siteng Huang$^{12}$ \qquad \\
    Mingyang Sun$^{12}$ \qquad
    Wei Zhao$^{1}$ \qquad 
    Zhaoxin Fan$^{3}$ \qquad
    Donglin Wang$^{1}$\textsuperscript{\Letter} \\
    \normalsize{$^{1}$Westlake University \qquad $^{2}$Zhejiang University \qquad}
    \vspace{-3pt} \\
    \normalsize{$^{3}$Beijing Advanced Innovation Center for Future Blockchain and Privacy Computing} \\
    {\tt\small \{gongzhefei, dingpengxiang, lyushangke, huangsiteng, wangdonglin\}@westlake.edu.cn} \\
    \small\textbf{{\url{https://carp-robot.github.io}}} \\
}

\newif\ifreview 
\newif\ifarxiv \newcommand{\arxiv}{\arxivtrue}
\newif\ifcamera 
\newif\ifrebuttal

\arxiv %

\pdfoutput=1
\documentclass[10pt,twocolumn,letterpaper]{article}
\ifreview \usepackage[review]{cvpr} \fi
\ifarxiv \usepackage[pagenumbers]{cvpr} \fi
\ifrebuttal \usepackage[rebuttal]{cvpr} \fi
\ifcamera \usepackage{cvpr} \fi

\usepackage{graphicx}	
\usepackage{amsmath}	
\usepackage{amssymb}	
\usepackage{booktabs}
\usepackage{times}
\usepackage{microtype}
\usepackage{epsfig}
\usepackage{caption}
\usepackage{float}
\usepackage{placeins}
\usepackage{stfloats}
\usepackage{enumitem}
\usepackage{tabularx}
\usepackage{xstring}
\usepackage{multirow}
\usepackage{colortbl}
\usepackage{color}
\usepackage{xspace}
\usepackage{url}
\usepackage{subcaption}
\usepackage[hang,flushmargin]{footmisc}
\usepackage{tikz}
\usepackage{algorithm}
\usepackage{algorithmic}
\usepackage{arydshln}
\usepackage{textcomp}
\usepackage{marvosym}
\definecolor{lightblue}{RGB}{171, 214, 226} 

\ifcamera \usepackage[accsupp]{axessibility} \fi

\ifarxiv  \fi

\newcommand{\R}[1]{{%
    \textbf{%
        \ifstrequal{#1}{1}{\textcolor{red}{R#1}}{%
        \ifstrequal{#1}{2}{\textcolor{blue}{R#1}}{%
        \ifstrequal{#1}{3}{\textcolor{magenta}{R#1}}{%
        \ifstrequal{#1}{4}{\textcolor{teal}{R#1}}{%
                           \textcolor{cyan}{R#1}%
        }}}}%
    }%
}}

\usepackage{xr-hyper}

\makeatletter
\newcommand*{\addFileDependency}[1]{
  \typeout{(#1)}
  \@addtofilelist{#1}
  \IfFileExists{#1}{}{\typeout{No file #1.}}
}

\makeatother
\newcommand*{\myexternaldocument}[1]{
    \externaldocument{#1}
    \addFileDependency{#1.tex}
    \addFileDependency{#1.aux}
}

\definecolor{linkorange}{rgb}{1.0, 0.5, 0.0}
\definecolor{cvprblue}{rgb}{0.21,0.49,0.74}
\definecolor{cvprpurple}{rgb}{0.5, 0.0, 0.5}
\usepackage[pagebackref,breaklinks,colorlinks,
    linkcolor=linkorange,
    citecolor=cvprpurple,
    urlcolor=linkorange
]{hyperref}

\usepackage[capitalize]{cleveref}
\crefname{section}{Sec.}{Secs.}
\crefname{table}{Table}{Tables}
\crefname{figure}{Fig.}{Figs.}

\ifarxiv \crefname{appendix}{App.}{Apps.}
\else \crefname{appendix}{Suppl.}{Suppls.} \fi

\unless\ifarxiv \myexternaldocument{_supplementary} \fi

\begin{document}

\title{\paperTitle}
\author{\authorBlock}
\maketitle

\begin{abstract}

In robotic visuomotor policy learning, diffusion-based models have achieved significant success in improving the accuracy of action trajectory generation compared to traditional autoregressive models. However, they suffer from inefficiency due to multiple denoising steps and limited flexibility from complex constraints.
In this paper, we introduce \textbf{C}oarse-to-Fine \textbf{A}uto\textbf{R}egressive \textbf{P}olicy (\textbf{CARP}), a novel paradigm for visuomotor policy learning that redefines the autoregressive action generation process as a coarse-to-fine, next-scale approach.
CARP decouples action generation into two stages: 
first, an action autoencoder learns multi-scale representations of the entire action sequence;
then, a GPT-style transformer refines the sequence prediction through a coarse-to-fine autoregressive process.
This straightforward and intuitive approach produces highly accurate and smooth actions, matching or even surpassing the performance of diffusion-based policies while maintaining efficiency on par with autoregressive policies.
We conduct extensive evaluations across diverse settings, including single-task and multi-task scenarios on state-based and image-based simulation benchmarks, as well as real-world tasks.
CARP achieves competitive success rates, with up to a 10\% improvement, and delivers \textbf{10×} faster inference compared to state-of-the-art policies, establishing a high-performance, efficient, and flexible paradigm for action generation in robotic tasks.

{
  \renewcommand{\thefootnote}
    {\fnsymbol{footnote}}
  \footnotetext[0]{$^{\dagger}$ Project lead. \textsuperscript{\Letter} Corresponding author.} 
  }
\end{abstract}

\section{Introduction}
\label{sec:introduction}

\begin{figure}[ht]
   \centering
   \includegraphics[width=1.0\linewidth]{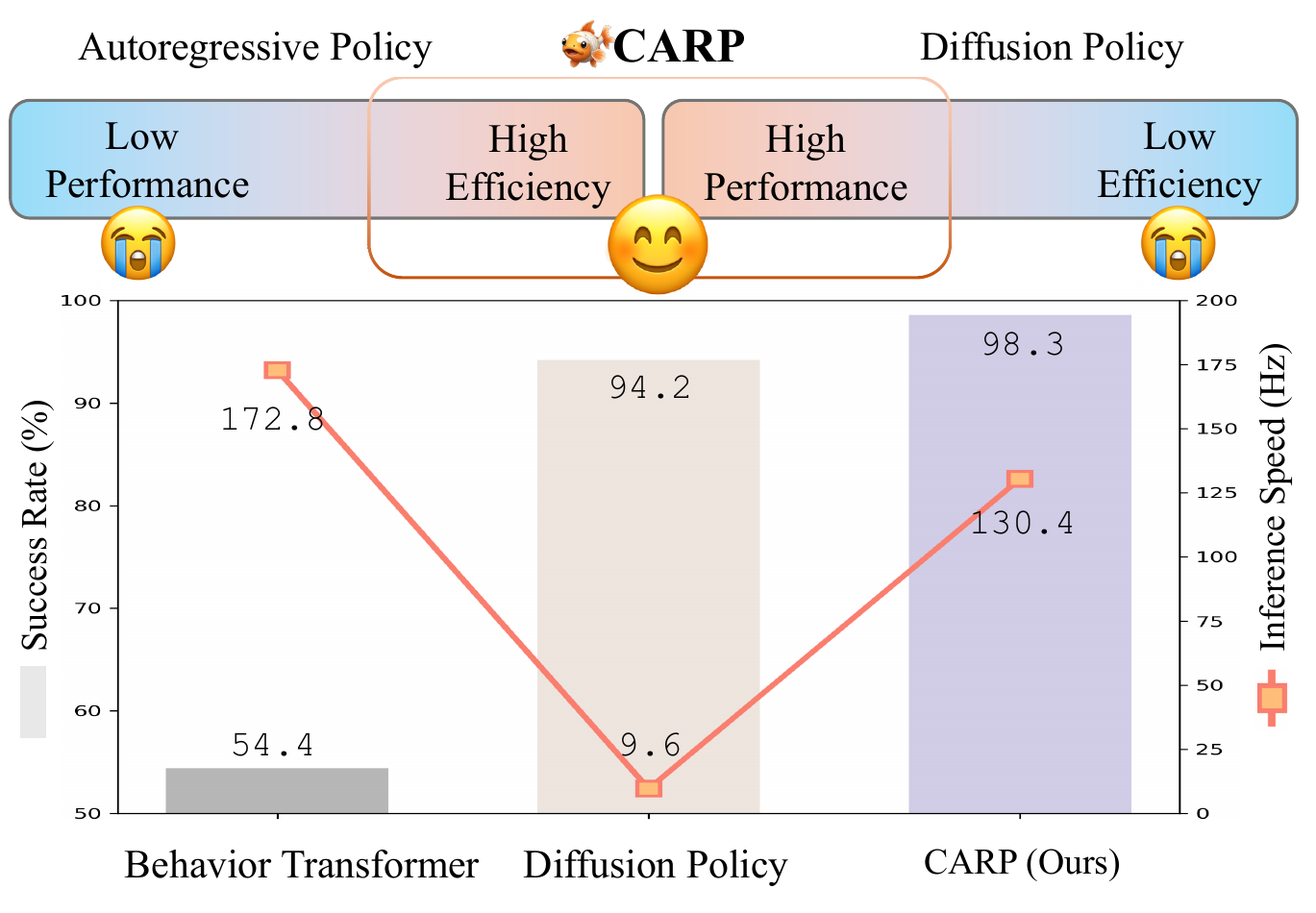}
   \vspace{-8mm}
   \caption{
   \textbf{Policy Comparison.} 
    The representative performance among Behavior Transformer~\cite{shafiullah2022behavior} served as an autoregressive policy, Diffusion Policy~\cite{chi2023diffusionpolicy}, and our approach in the state-based Robomimic Square task experiment. 
    CARP demonstrates an effective balance between task performance and inference efficiency.
    }
   \label{fig:teaser}
   \vspace{-4mm}
\end{figure}

\begin{figure*}[ht]
  \centering
  \begin{subfigure}{0.25\linewidth}
    \centering
    \includegraphics[height=3cm]{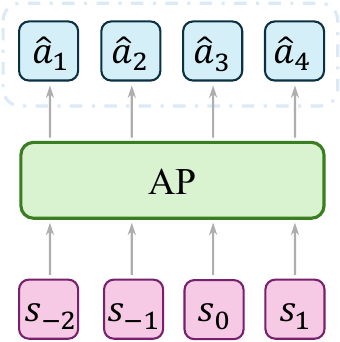}
    \caption{Autoregressive Policy}
    \label{fig:intro_ar}
  \end{subfigure}
  \hfill
  \begin{subfigure}{0.25\linewidth}
    \centering
    \includegraphics[height=3cm]{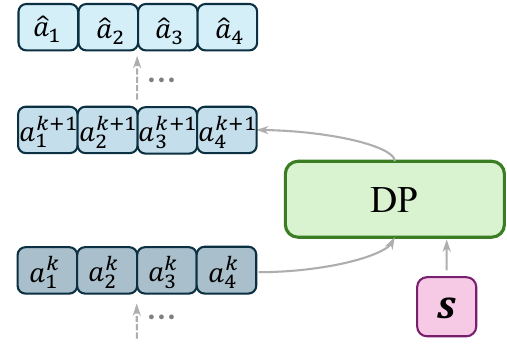}
    \caption{Diffusion Policy}
    \label{fig:intro_dp}
  \end{subfigure}
  \hfill
  \begin{subfigure}{0.46\linewidth}
    \centering
    \includegraphics[height=3cm]{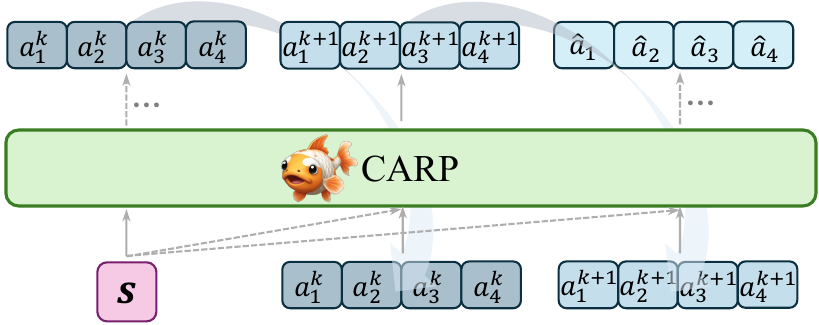}
    \caption{CARP (Ours)}
    \label{fig:intro_carp}
  \end{subfigure}
  \vspace{-2.5mm}
  \caption{
  \textbf{Structure of Current Policies.}
  $\hat{\boldsymbol{a}}$ is the predicted action, $\boldsymbol{a}^k$ denotes the refining action at step $k$, $\boldsymbol{s}$ is the historical condition.
  a) Autoregressive Policy predicts the action step-by-step in the next-token paradigm.
  b) Diffusion Policy models the noise process used to refine the action sequence.
  c) CARP refines action sequence predictions autoregressively from coarse to fine granularity.}
  \label{fig:compare_3}
  \vspace{-2.5mm}
\end{figure*}

Policy learning from demonstrations, formulated as the supervised regression task of mapping observations to actions, has proven highly effective across 
a wide range of 
robotic tasks, even in its simplest form. 
Replacing the traditional policies with generative models, particularly those stemming from the vision community, has opened new avenues for improving performance, enabling robots to achieve the high precision required for complex tasks in robotic scenarios.

Existing approaches have explored different generative modeling techniques to address challenges in visuomotor policy learning. 
Autoregressive Modeling (AM) \cite{shafiullah2022behavior, zhao2023learning, cui2022play, lee2024behavior} provides a straightforward and efficient out-of-the-box solution, benefitting from its scalability, flexibility, and mature exploration at lower computation requirements. 
However, AM's next-token prediction paradigm often fails to capture long-range dependencies, global structure, and temporal coherence~\cite{kilian2024computational}, leading to poor performance, which is essential for many robotic tasks. 
Recently, Diffusion Modeling (DM) \cite{chi2023diffusionpolicy, wang2024sparse, reuss2024multimodal} has emerged as a promising alternative, bridging the precision gap in AM by modeling the gradient of the action score function to learn multimodal distributions. Nevertheless, DM requires multiple steps of sequential denoising, making it computationally prohibitive for robotic tasks, especially for robotic tasks requiring efficient real-time inference in on-board compute-constrained environments.
Additionally, DDPM~\cite{ho2020denoising}’s rigid generative process lacks flexibility and adaptability for tasks with long-term dependencies, often leading to cumulative errors and reduced robustness over extended time spans, limiting its use in dynamic settings.

Both AM and DM have their respective advantages and limitations, which are often orthogonal and difficult to balance in practical applications. In this work, we aim to resolve this trade-off by introducing a novel generative paradigm for robot visuomotor policy learning that predicts entire action sequences from a \textit{coarse-to-fine} granularity in a \textit{next-scale} prediction framework. 
This approach allows our model to achieve performance levels comparable to DM while retaining AM’s inference efficiency and implementation flexibility.

Our primary contribution is the introduction of a \textbf{C}oarse-to-Fine \textbf{A}uto\textbf{R}egressive \textbf{P}olicy (\textbf{CARP}), a hybrid framework that combines AM’s efficiency with DM’s high performance to meet the demands of real-world robotic manipulation. 
Specifically, our main contributions are as follows:
\begin{itemize}
    \item \textbf{Multi-Scale action tokenization:} We propose a multi-scale tokenization method for action sequences that captures the global structure and maintains temporal locality, effectively addressing AM’s myopic limitations.
    \item \textbf{Coarse-to-Fine autoregressive prediction:} This mechanism refines action sequences in the latent space using Cross-Entropy loss with relaxed Markovian assumptions during iterations, achieving DM-like performance with high efficiency and comparable multi-modal behavior.
    \item \textbf{Comprehensive sim \& real experiments:} Extensive experiments demonstrate CARP’s effectiveness in both simulated and real-world robotic manipulation tasks.
\end{itemize}
In summary, we present CARP, a novel visuomotor policy framework that synergizes the strengths of AM and DM, offering high performance, efficiency, and flexibility. 

\begin{figure*}[ht]
  \centering
  \begin{subfigure}{0.44\textwidth} 
    \centering
    \includegraphics[height=8cm, width=\linewidth, keepaspectratio]{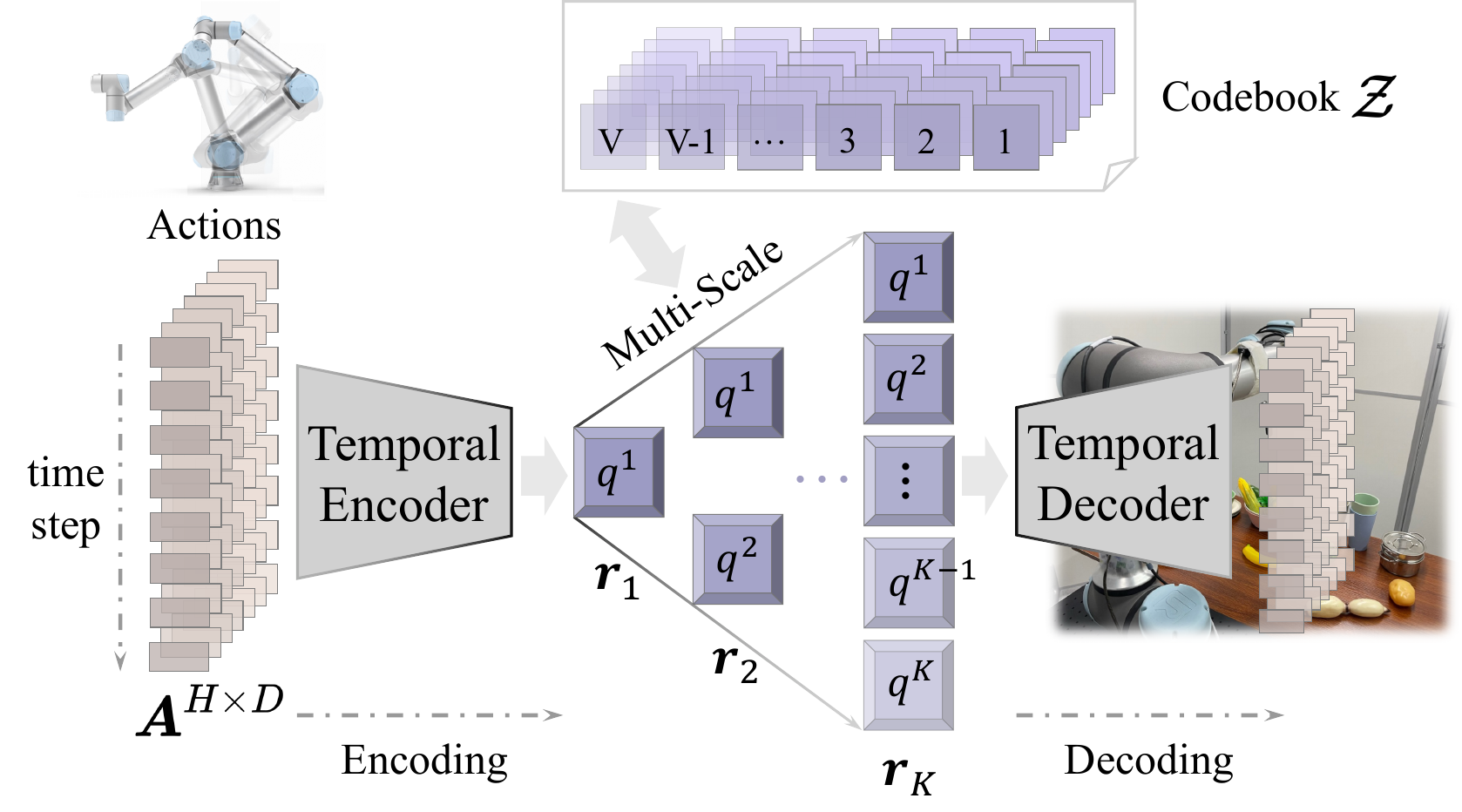} 
    \caption{Multi-Scale Action Tokenization}
    \label{fig:carp_vqvae}
  \end{subfigure}
  \hspace{0.2mm}
  \begin{tikzpicture}
    \draw[dashed] (0,0) -- (0,5.25);
  \end{tikzpicture}
  \begin{subfigure}{0.54\textwidth}
    \centering
    \includegraphics[height=8cm, width=\linewidth, keepaspectratio]{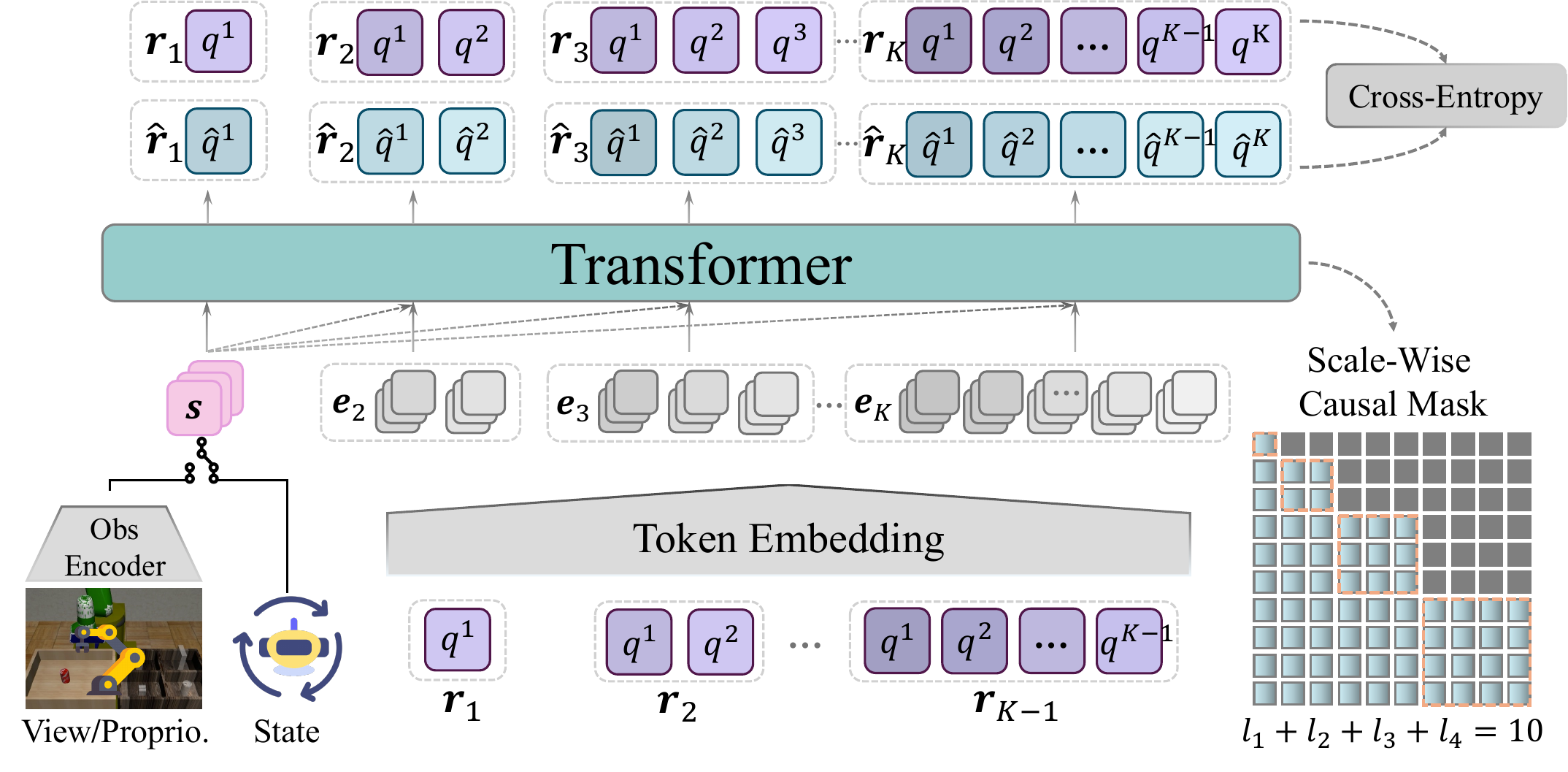}
    \caption{Coarse-to-Fine Autoregressive Prediction }
    \label{fig:carp_ar}
  \end{subfigure}
  \vspace{-7mm}
  \caption{\textbf{Overview of the Two Stages of CARP.} 
  a) A multi-scale action autoencoder extracts token maps $\boldsymbol{r}_1, \boldsymbol{r}_2, \dots, \boldsymbol{r}_K$ to represent the action sequence at different scales, trained using the standard VQVAE loss.
  b) The autoregressive prediction is reformulated as a \textit{coarse-to-fine}, \textit{next-scale} paradigm.
  The sequence is progressively refined from coarse token map $\boldsymbol{r}_1$ to finer granularity token map $\boldsymbol{r}_K$, where each $\boldsymbol{r}_k$ contains $l_k$ tokens.
  An attention mask ensures that each $\boldsymbol{r}_k$ attends only to the preceding $\boldsymbol{r}_{1:k-1}$ during training.
  A standard Cross-Entropy loss is used for training. 
  During inference, the token maps $\boldsymbol{r}_{1:K}$ are collectively decoded into continuous actions for execution.
  }
  \label{fig:carp}
  \vspace{-2mm}
\end{figure*}

\section{Background}
\label{sec:background}

We start by providing the background of our approach, focusing on three key components: problem formulation, conventional autoregressive policies, and diffusion-based policies.

\subsection{Problem Formulation}
Problem formulation will consider a task $\mathcal{T}$, where there are $N$ expert demonstrations $\{\tau_{i}\}_{i=1}^{N}$. Each demonstration $\tau_{i}$ is a sequence of state-action pairs. We formulate robot imitation learning as an action sequence prediction problem \cite{Rpt2023,chi2023diffusionpolicy,wang2024sparse}, training a model to minimize the error between the predicted future actions conditioned on historical states and the ground truth actions.
Specifically, imitation learning minimize the behavior cloning loss $\mathcal{L}_{bc}$ formulated as
\vspace{-2.5mm}
\begin{equation}
\mathcal{L}_{bc} = \mathbb{E}_{\boldsymbol{s}, \boldsymbol{a} \sim \mathcal{T}} 
\left[\sum_{t=0}^{T} \mathcal{L}\left(\pi_{\boldsymbol{\theta}}(\hat{\boldsymbol{a}}_H | \boldsymbol{s}_O), \boldsymbol{a}_H\right)\right],
\label{eq:bc-loss}
\vspace{-2.5mm}
\end{equation}
where $\boldsymbol{a}$ represents the action, $\boldsymbol{s}$ denotes the state or observation according to the specific task description, $t$ is the current time step, $H$ is the prediction horizon, and $O$ is the historical horizon. For notational simplicity, we denote the action sequence $\boldsymbol{a}_{t : t + H - 1}$ as $\boldsymbol{a}_H$ and the state sequence $\boldsymbol{s}_{t - O + 1 : t}$ as $\boldsymbol{s}_O$. Here, $\mathcal{L}_{bc}$ represents a supervised action prediction loss, such as mean squared error or negative log-likelihood, $T$ is the length of the demonstration, and $\theta$ represents the learnable parameters of the policy network $\pi_{\boldsymbol{\theta}}$.

\subsection{Autoregressive Policy} 
Autoregressive policies leverage the efficiency and flexibility of autoregressive models (GPT-style Decoders). Recent advancements, such as action chunking~\cite{shafiullah2022behavior, zhao2023learning}, redefine this paradigm as a multi-token, one-pass prediction method 
(see Suppl. C for further details). 
We refer to this approach as Autoregressive Policy (AP).
In this approach, the \textit{next-token} posits the probability of observing the current action $\boldsymbol{a}_t$ depends solely on its previous states $\boldsymbol{s}_O$, which allows for the factorization of the likelihood of sequence with length $H$ as 
\vspace{-2.5mm}
\begin{equation}
p\left(\boldsymbol{a}_t,\boldsymbol{a}_{t+1},...,\boldsymbol{a}_{t+H-1}\right) = \prod^{t+H-1}_{k=t}p\left(\boldsymbol{a}_k|\boldsymbol{s}_O\right).
\label{eq:ar_prediction}
\vspace{-2.5mm}
\end{equation}
However, it introduces several issues that hinder its performance. The linear, step-by-step unidirectional dependency of action prediction may overlook the global structure~\cite{kilian2024computational}, making it challenging to capture long-range dependencies and holistic coherence or temporal locality~\cite{janner2022diffuser} in complex scenes or sequences, which restricts their generalizability in tasks requiring bidirectional reasoning, as shown in \cref{fig:intro_ar}. For example, it cannot predict the former actions given the near-terminal state without backward reasoning.

\subsection{Diffusion Policy}
Diffusion-based policies~\cite{chi2023diffusionpolicy} utilize Denoising Diffusion Probabilistic Models~\cite{ho2020denoising} to approximate the conditional distribution $p(\boldsymbol{a}_{H}|\boldsymbol{s}_{O})$ instead of the joint distribution~\cite{janner2022diffuser} $p(\boldsymbol{a}_{H},\boldsymbol{s}_{O})$, through modeling the noise during the denoising process from Gaussian noise to noise-free output as 
\vspace{-2.5mm}
\begin{equation}
    \boldsymbol{a}_{H}^{k+1} = \alpha(\boldsymbol{a}_{H}^{k} - \gamma\epsilon_\theta(\boldsymbol{s}_{O},\boldsymbol{a}_{H}^{k},k) + \mathcal{N}),
\label{eq:dp_langevin}
\vspace{-2.5mm}
\end{equation}
where $\epsilon_\theta$ is a learnable noise network, $k$ is the current denoising step, $\mathcal{N}$ is Gaussian noise, and $\alpha$, $\gamma$ are hyper-parameters.

Diffusion-based policies show impressive performance in robotic manipulation tasks due to their action generation which gradually refines from random samples which we can abstract as a \textit{coarse-to-fine} process. 
This kind of process alike human movement or natural thinking flows in line with human intuition. 
However, they suffer from poor convergence due to their design. To address this, multiple denoising steps are required, constrained by the Markovian assumption, where $\boldsymbol{a}_{H}^{k+1}$ depends only on the previous step $\boldsymbol{a}_{H}^k$ 
(see \cref{fig:intro_dp})
, which leads to significant runtime inefficiency.
Moreover, they lack the ability to leverage generation context effectively, hindering scalability to complex scenes~\cite{gu2024dart}.

\section{Method}
\label{sec:method}

To address the limitations of existing methods, we propose \textbf{C}oarse-to-Fine \textbf{A}uto\textbf{R}egressive \textbf{P}olicy (\textbf{CARP}), a novel visuomotor policy framework that combines the high performance of recent diffusion-based policies with the inference efficiency and flexibility of traditional autoregressive policies.
CARP achieves these advantages through a redesigned autoregressive modeling strategy. Specifically, we shift from conventional \textit{next-token} prediction to a \textit{coarse-to-fine}, \textit{next-scale} prediction approach, using \textit{multi-scale} action representations. 
The training process of CARP consists of two stages: \textit{multi-scale} action tokenization and \textit{coarse-to-fine} autoregressive prediction, as detailed in \cref{fig:carp}.

In this section, we first explain the construction of \textit{multi-scale} action token maps, followed by the \textit{coarse-to-fine} autoregressive prediction approach for action generation. Key implementation details are provided at the end.

\subsection{Multi-Scale Action Tokenization}
\label{sec:method_vqvae}
Instead of focusing on individual action steps, we extract representations at multiple scales across the entire action sequence. 
We propose a novel \textit{multi-scale} action quantization autoencoder that encodes a sequence of actions into $K$ discrete token maps, $\boldsymbol{R} = (\boldsymbol{r}_1, \boldsymbol{r}_2, \dots, \boldsymbol{r}_K)$, which are used for
training and inference. Our approach builds on the VQVAE architecture~\cite{van2017neural}, incorporating a modified \textit{multi-scale} quantization layer~\cite{tian2024visual} 
to enable hierarchical encoding.

\textbf{Encoder and Decoder.}
As illustrated in \cref{fig:carp_vqvae}, the actions are first organized into an action sequence $\boldsymbol{A}^{H \times D}$, where $H$ denotes the prediction horizon and $D$ represents the dimensionality of the action $\boldsymbol{a}$. Given that each dimension in the action space is orthogonal to the others, and that there exists a natural temporal dependency within an action sequence, we employ a 1D temporal convolutional network (1D-CNN) along the time dimension, similar to the architecture used in~\cite{janner2022diffuser}, for both the encoder $\mathcal{E}(\cdot)$ and decoder $\mathcal{D}(\cdot)$. 
Let $\boldsymbol{F} = \mathcal{E}(\boldsymbol{A}) \in \mathbb{R}^{L \times C}$, where $L$ denotes the compressed length of the temporal dimension with $L \leq H$, and $C$ represents the dimensionality of the feature map $\boldsymbol{F}$.

\textbf{Quantization.}
We introduce a quantizer with a learnable codebook $\mathcal{Z} \in \mathbb{R}^{V \times C}$, containing $V$ code vectors, each of dimension $C$.
The quantization process (lines 4-11 in \cref{alg:act_vqvae}) generates the action sequence representation iteratively across multiple scales. 
At scale $k$, it produces action token map $\boldsymbol{r}_k$ to represent the sequence, which consists of $l_k$ tokens $q$. In our implementation, we set $l_k = k$. 
The function $\operatorname{Lookup}(\mathcal{Z}, v)$ retrieves the $v$-th code vector from $\mathcal{Z}$. The quantization function is defined by $\boldsymbol{r} = \mathcal{Q}(\boldsymbol{F})$ as 
\vspace{-2mm}
\begin{equation}
    q = \operatorname*{arg\,min}_{v \in [V]}
    \operatorname{dist}\left(\operatorname{Lookup}(\mathcal{Z}, v), \boldsymbol{f} \right) \in [V],
\label{eq:act_quantizer}
\vspace{-2mm}
\end{equation}
where token $q$ represents the nearest vector in $\mathcal{Z}$ for a given feature vector $\boldsymbol{f} \in \mathbb{R}^{1 \times C}$ in the feature map $\boldsymbol{F}$, according to a distance function $\operatorname{dist}(\cdot)$, such as Euclidean, cosine, etc.

Specifically, we adopt a residual-style design~\cite{lee2022autoregressive, tian2024visual} for feature maps $\boldsymbol{F}$ and $\hat{\boldsymbol{F}}$, as shown in lines 4-11 of \cref{alg:act_vqvae}. 
This design ensures that each finer-scale representation $\boldsymbol{r}_k$ depends only on its coarser-scale predecessors $(\boldsymbol{r}_1, \boldsymbol{r}_2, \dots, \boldsymbol{r}_{k-1})$, facilitating a \textit{multi-scale} representation. 
A shared codebook $\mathcal{Z}$ is used across all scales, ensuring that \textit{multi-scale} token maps $\boldsymbol{R} = (\boldsymbol{r}_1, \boldsymbol{r}_2, \dots, \boldsymbol{r}_K)$, where $K$ is the number of scales, are drawn from a consistent vocabulary $[V]$.
To preserve information during upsampling, we employ $K$ additional 1D convolutional layers $\{\phi_k\}_{k=1}^K$~\cite{tian2024visual}, as illustrated in lines 9-10 of \cref{alg:act_vqvae}.

\textbf{Loss.}
The final approximation $\hat{\boldsymbol{F}}$ of the original feature map $\boldsymbol{F}$ is composed residually of the \textit{multi-scale} representations derived from the codebook $\mathcal{Z}$, based on all token maps $\{\boldsymbol{r}_k\}_{k=1}^K$. 
The reconstructed action sequence is then obtained through $\hat{\boldsymbol{A}} = \mathcal{D}(\hat{\boldsymbol{F}})$.
To train the quantized autoencoder, a typical VQVAE~\cite{van2017neural} loss $\mathcal{L}$ is minimized as
\vspace{-2mm}
\begin{equation}
    \mathcal{L} = \underbrace{\|\boldsymbol{A} - \hat{\boldsymbol{A}}\|_2}_{\mathcal{L}_{\text{recon}}} + \underbrace{\|\text{sg}(\boldsymbol{F})-\hat{\boldsymbol{F}}\|_2}_{\mathcal{L}_{\text{quant}}} + \underbrace{\|\boldsymbol{F}-\text{sg}(\hat{\boldsymbol{F}})\|_2}_{\mathcal{L}_{\text{commit}}},
\label{eq:act_vqvae_loss}
\vspace{-2.5mm}
\end{equation}
where $\mathcal{L}_{\text{recon}}$ minimizes the difference between the original action sequence $\boldsymbol{A}$ and the reconstruction $\hat{\boldsymbol{A}}$. $\text{sg}(\cdot)$ denotes stop-gradient. $\mathcal{L}_{\text{quant}}$ aligns the quantized feature map $\hat{\boldsymbol{F}}$ with the original $\boldsymbol{F}$, and $\mathcal{L}_{\text{commit}}$ encourages the encoder to commit to codebook entries, preventing codebook collapse.
For $\mathcal{L}_{\text{quant}}$ and $\mathcal{L}_{\text{commit}}$, we calculate every residual calculating moment of each scale $\boldsymbol{r}_k$ as in \cref{alg:act_vqvae}. 
After the training process, the autoencoder $\{\mathcal{E}, \mathcal{Q}, \mathcal{D}\}$ tokenizes actions for subsequent \textit{coarse-to-fine} autoregressive modeling.

\textbf{Discussion.}
The tokenization strategy described above allows the \textit{multi-scale} tokens $\boldsymbol{R}$, extracted from the action sequence $\boldsymbol{A} \in \mathbb{R}^{H \times D}$ via temporal 1D convolutions, to inherently preserve temporal locality~\cite{janner2022diffuser}. Additionally, the hierarchical extraction captures the global structure, enabling the model to treat the action sequence as a unified entity.

Unlike traditional autoregressive policies that predict each action token independently (see \cref{eq:ar_prediction} and \cref{fig:intro_ar}), CARP leverages dual capabilities to capture both local temporal dependencies and global features across the entire action sequence. This approach produces smoother transitions and more stable action sequences. 
Through \textit{multi-scale} encoding, CARP overcomes the short-sighted limitations of conventional autoregressive models, yielding more robust, coherent, and precise behaviors over extended time horizons.

\begin{algorithm}[ht]
\caption{\small Multi-Scale Action VQVAE}
\label{alg:act_vqvae}
\begin{algorithmic}[1]
    \small
    \STATE \textbf{Inputs:} Action sequence  $\boldsymbol{A}$
    \STATE \textbf{Hyperparameters:} Number of scales $K$, length of each scale $(l_k)_{k=1}^{K}$ , length of feature map's temporal dimension $L$
    \STATE \textbf{Initialize:} $\boldsymbol{F} \gets \mathcal{E}(\boldsymbol{A})$, $\hat{\boldsymbol{F}}\gets 0$, $R\gets[]$
    \FOR{$k = 1$ \textbf{to} $K$}
        \STATE $\boldsymbol{r}_k \gets \mathcal{Q}(\text{Interpolate}(\boldsymbol{F}, l_k))$
        \STATE $\boldsymbol{R} \gets \boldsymbol{R} \cup \{\boldsymbol{r}_k\}$
        \STATE $\boldsymbol{Z}_k \gets \text{Lookup}(\mathcal{Z}, \boldsymbol{r}_k)$
        \STATE $\boldsymbol{Z}_k \gets \text{Interpolate}(\boldsymbol{Z}_k, L)$
        \STATE $\boldsymbol{F} \gets \boldsymbol{F} - \phi_k(\boldsymbol{Z}_k)$
        \STATE $\hat{\boldsymbol{F}} \gets \hat{\boldsymbol{F}} + \phi_k(\boldsymbol{Z}_k)$
    \ENDFOR
    \STATE $\hat{\boldsymbol{A}} \gets \mathcal{D}(\hat{\boldsymbol{F}})$
    \STATE \textbf{Return:} Multi-scale token maps $\boldsymbol{R}$, reconstructed action sequence $\hat{\boldsymbol{A}}$
\end{algorithmic}
\end{algorithm}

\subsection{Coarse-to-Fine Autoregressive Prediction} 
\label{sec:method_ar}
Using \textit{multi-scale} action sequence representations, we shift from traditional \textit{next-token} prediction to a \textit{next-scale} prediction approach, progressing from coarse to fine granularity.

\textbf{Prediction.}
Given the \textit{multi-scale} representation token maps $(\boldsymbol{r}_1, \boldsymbol{r}_2, \dots, \boldsymbol{r}_K)$ produced by the trained autoencoder, where each $\boldsymbol{r}_k$ encodes the same action sequence $\boldsymbol{A}$ at a different scale, the autoregressive likelihood is formulated as
\vspace{-5mm}
\begin{equation}
    p(\boldsymbol{r}_1, \boldsymbol{r}_2, \dots, \boldsymbol{r}_K) = \prod_{k=1}^{K} p(\boldsymbol{r}_k \mid \boldsymbol{r}_1, \boldsymbol{r}_2, \dots, \boldsymbol{r}_{k-1};\boldsymbol{s}_O),  
\label{eq:cap_predict}
\vspace{-1mm}
\end{equation}
where each autoregressive unit $\boldsymbol{r}_k \in [V]^{l_k}$ is the token map at scale $k$ containing $l_k$ tokens $q$. 
The predix sequence $(\boldsymbol{r}_1, \boldsymbol{r}_2, \dots, \boldsymbol{r}_{k-1})$ is served as the condition for $\boldsymbol{r}_k$, accompanying with the historical state sequence $\boldsymbol{s}_O$. This kind of \textit{next-scale} prediction methodology is what we define as \textit{coarse-to-fine} autoregressive prediction in CARP.

Due to the residual-style quantization, during autoregressive prediction, we first embed the previous scale token map $\boldsymbol{r}_{k-1}$ and then reconstruct the next scale feature map $\boldsymbol{e}_{k}$, as shown in \cref{fig:carp_ar}. This feature map $\boldsymbol{e}_{k}$ is then used as input for predicting the next scale token map $\boldsymbol{r}_k$.
The token maps $\boldsymbol{r}_{1:K}$ are decoded by the \textit{multi-scale} autoencoder into continuous actions, following the residual-style process.

\textbf{Loss.}
During the $k$-th autoregressive step, all distributions over the $l_k$ tokens in $\boldsymbol{r}_k$ will be generated in parallel, with the \textit{coarse-to-fine} dependency ensured by a block-wise causal attention mask~\cite{tian2024visual}. 
To optimize the autoregressive model, we utilize the standard Cross-Entropy loss to capture the difference between the predicted token map $\hat{\boldsymbol{r}}$ and the token map $\boldsymbol{r}$ from the ground truth action sequence as
\vspace{-2.5mm}
\begin{equation}
    \mathcal{L}_{\text{Cross-Entropy}} = \sum_{k=1}^K \sum_{i=1}^{l_k} \left[- \sum_{v=1}^V \boldsymbol{r}_k^{i, v} \log \hat{\boldsymbol{r}}_k^{i, v}\right],
\label{eq:cap_ar_loss}
\vspace{-2.5mm}
\end{equation}
where $l_k$ is the length of each scale, 
$V$ is the size of the action dictionary $\mathcal{Z}$, 
and $K$ is the number of scales.

\textbf{Discussion.} 
CARP models the entire trajectory holistically, progressively refining actions from high-level intentions to fine-grained details (see \cref{fig:intro_carp}). 
This hierarchical paradigm more closely resembles natural human behavior, where movements are guided by overarching goals and gradually refined, rather than being planned step by step.

Instead of the commonly used MSE loss~\cite{zhao2023learning,torabi2018behavioral,bojarski2016end}, CARP employs Cross-Entropy loss (\cref{eq:cap_ar_loss}) to preserve multimodality naturally, as MSE tends to enforce a unimodal distribution that can be detrimental to robotic tasks~\cite{shafiullah2022behavior}.
CARP’s iterative refinement process resembles the denoising steps in diffusion models to achieve high accuracy.

Furthermore, our approach models actions directly rather than modeling noise, facilitating faster prediction convergence in low-dimensional manifolds~\cite{lu2024manicm, ze20243d}. 
By operating in the latent space with action sequence tokens, CARP mitigates trajectory anomalies that can disrupt prediction, unlike methods that work on raw actions~\cite{chi2023diffusionpolicy}. 
This latent-space representation allows CARP to focus on the essential components of actions, resulting in smoother and more efficient predictions.
In contrast, traditional diffusion models rely on Markovian processes, which compress all information into progressively noisier inputs from previous levels, often hindering efficient learning and requiring more inference steps~\cite{gu2024dart} (see \cref{eq:dp_langevin} and \cref{fig:intro_dp}). CARP relaxes this constraint by allowing each scale $\boldsymbol{r}_k$ to depend on all prior scales $\boldsymbol{r}_{1:k-1}$ instead of the only previous scale $\boldsymbol{r}_{k-1}$. 
This structure enables CARP to generate high-quality trajectories with significantly fewer steps, demonstrating superior efficiency.

\subsection{Implementation Details}
Here, we present the key implementation details of CARP.

\textbf{Tokenization.}
Due to the disentangling of the action prediction, imprecise \textit{coarse-to-fine} action representations can decrease the upper limit of model performance. 
Considering the discontinuity of the most representation for rotation space like Euler angle or quaternion will increase the unstable training process, we utilize rotation6d~\cite{zhou2019continuity} to get stable \textit{multi-scale} tokens. 
And for the distance function $\operatorname{dist}(\cdot)$, we use cosine similarity rather than Euclidean distance which is the cause of unstable training based on our observations.
In practice, due to the orthogonality between dimensions, we use a separate VQVAE~\cite{tian2024visual} for each dimension of the action space, to gain stable training.
All convolutions we used in CARP are 1D to capture the time-dimension features.

\textbf{Autoregressive.} We adopt the architecture of standard decoder-only transformers akin to GPT-2~\cite{tian2024visual, chen2023pixart}. The state $\boldsymbol{s}_{O}$ is used to generate the initial coarse-scale token map and then is utilized as adaptive normalization~\cite{park2019semantic} for the subsequent predictions. 
During training, we observe that incorporating Exponential Moving Average (EMA)~\cite{haynes2012exponential} enhances both training stability and performance, yielding a 4-5 \% improvement, consistent with findings in~\cite{chi2023diffusionpolicy}.
During the inference, kv-caching can be used and no mask is needed (for further inference details, refer to Suppl. B).

\section{Experiment}
\label{sec:experiment}

In this section, we evaluate CARP on diverse robotics tasks, including state-based and image-based benchmarks in single-task and multi-task settings.
CARP's performance is assessed in terms of task success rate, inference speed, and model parameter scale.
Additionally, we validate CARP's practical effectiveness by deploying it on real-world tasks using UR5e and Franka robotic arms, comparing its performance against state-of-the-art diffusion-based policies.
Our experiments are structured to address the following key research questions
(see Suppl. H for experimental implementation details):
\begin{itemize}
    \item \textbf{RQ1}: Can CARP achieve accuracy and robustness comparable to current state-of-the-art diffusion-based policies?
    \item \textbf{RQ2}: Does CARP maintain high inference efficiency, characteristic of autoregressive models?
    \item \textbf{RQ3}: Does CARP leverage the flexibility benefits of a GPT-style architecture?
\end{itemize}

\vspace{-2mm}
\begin{figure}[ht]
   \centering
   \includegraphics[width=1.0\linewidth]{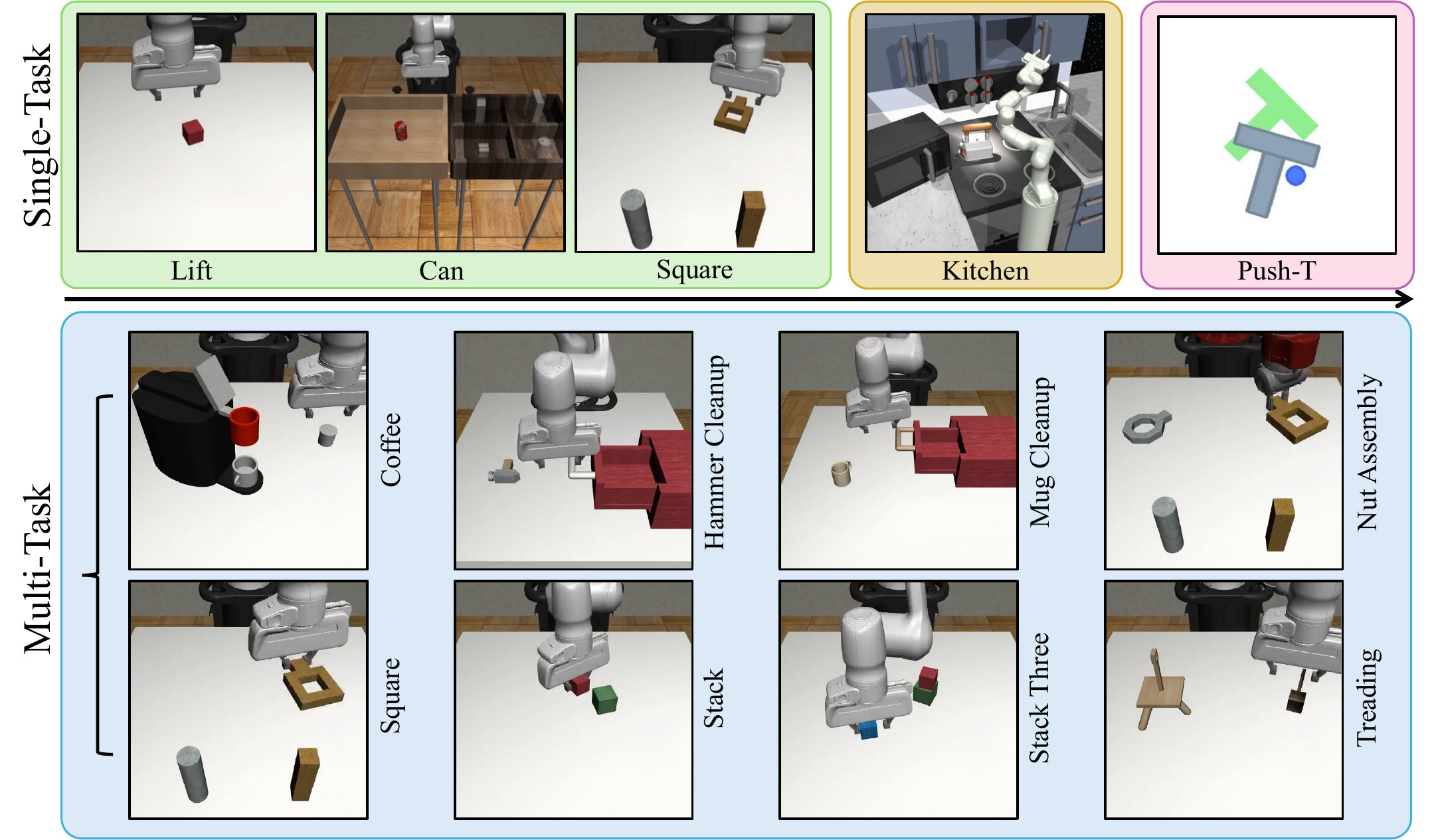}
   \vspace{-6mm}
   \caption{
   \textbf{Simulation Tasks Visualization.}  
    In the single-task setting, we evaluate Lift, Can, and Square, ordered by increasing difficulty, along with the Kitchen task for long-horizon evaluation and Push-T task for assessing multi-modal behavior.
    In the multi-task setting, we consider eight tasks: Coffee, Hammer Cleanup, Mug Cleanup, Nut Assembly, Square, Stack, Stack Three, and Threading, arranged from left to right and top to bottom.
   }
   \vspace{-2.5mm}
   \label{fig:sim_task_vis}
\end{figure}

\subsection{Evaluation on Single-Task Simulations}
\label{sec:exp_sim}
We begin by evaluating CARP on a set of standard simulated tasks commonly used to benchmark diffusion-based policies, aiming to assess whether it can achieve comparable performance while significantly improving inference efficiency.

\begin{table}[ht]
  \centering
  \small
  \begin{tabular}{@{}lccccc@{}}
    \toprule
    Policy & Lift & Can & Square & Params/M & Speed/s \\
    \midrule
    BET~\cite{shafiullah2022behavior} & 0.96 & 0.88 & 0.54 & \textbf{0.27} & \textbf{2.12} \\
    DP-C~\cite{chi2023diffusionpolicy} & \textbf{1.00} & 0.94 & \textbf{0.94} & 65.88 & 35.21\\
    DP-T~\cite{chi2023diffusionpolicy} & \textbf{1.00} & \textbf{1.00} & 0.88 & 8.97 & 37.83\\
    \rowcolor{cyan!20}
    CARP (Ours) & \textbf{1.00} & \textbf{1.00} & \textbf{0.98} & \textbf{0.65} & \textbf{3.07} \\
    \bottomrule
  \end{tabular}
  \vspace{-2.5mm}
  \caption{
  \textbf{State-Based Simulation Results (State Policy).} 
    We report the average success rate of the top 3 checkpoints, along with model parameter scales and inference time for generating 400 actions. CARP significantly outperforms BET and achieves competitive performance with state-of-the-art diffusion models, while also surpassing DP in terms of model size and inference speed.}
  \label{tab:state_robomimic}
\end{table}

\begin{table}[ht]
  \centering
  \small
  \begin{tabular}{@{}lccccc@{}}
    \toprule
    Policy & Lift & Can & Square & Params/M & Speed/s \\
    \midrule
    IBC~\cite{pmlr-v164-florence22a} & 0.72 & 0.02 & 0.00 & \textbf{3.44} & 32.35 \\
    DP-C~\cite{chi2023diffusionpolicy} & \textbf{1.00} & 0.97 & \textbf{0.92} & 255.61 & 47.37 \\
    DP-T~\cite{chi2023diffusionpolicy} & \textbf{1.00} & \textbf{0.98} & 0.86 & 9.01 & 45.12 \\
    \rowcolor{cyan!20}
    CARP (Ours) & \textbf{1.00} & \textbf{0.98} & \textbf{0.88} & \textbf{7.58} & \textbf{4.83}\\
    \bottomrule
  \end{tabular}
  \vspace{-2.5mm}
  \caption{
  \textbf{Image-Based Simulation Results (Visual Policy).} 
  Results show that CARP consistently balances high performance and high inference efficiency. 
  We highlight our results in \textcolor{lightblue}{light-blue}.}
  \label{tab:image_robomimic}
\end{table}

\textbf{Experimental Setup.} 
We use the Robomimic~\cite{robomimic2021} benchmark suite, which is widely adopted for evaluating diffusion-based policies~\cite{chi2023diffusionpolicy,wang2024sparse,prasad2024consistency}. 
We evaluate CARP on both state-based and image-based datasets collected from experts, with each task containing 200 demonstrations.
The selected tasks are standard in single-task evaluations, as shown in \cref{fig:sim_task_vis}, with each task predict max 400 actions to execute the task.
For evaluating the ability to learn multiple long-horizon tasks, we utilize the Franka Kitchen environment~\cite{gupta2019relay}, which contains 7 objects for interaction and comes with a human demonstration dataset containing 566 trajectories, each completing 4 tasks in arbitrary order (see Suppl. M for more visualization).
To assess multi-modal behavior, we use the Push-T task from IBC~\cite{pmlr-v164-florence22a}, which involves contact-rich dynamics to push a T-shaped block along multiple paths. The state-based observation includes 9$\times$2D keypoints from the block’s ground-truth pose and proprioception of the end-effector. We train with 200 expert demonstrations.

\textbf{Baselines.}
We compare CARP with previous autoregressive policies and recent diffusion policies. 
Behavior Transformer (BET)~\cite{shafiullah2022behavior} is an autoregressive policy with action discretization and correction mechanisms, similar to offset-based prediction.
We further extend BET with action chunking to improve performance, as shown in \cref{fig:intro_ar}.
Implicit Behavior Cloning (IBC)~\cite{pmlr-v164-florence22a} utilizes energy-based models for supervised robotic behavior learning.
Diffusion Policy (DP)~\cite{chi2023diffusionpolicy} combines a denoising process with action prediction, implemented in two variants: CNN-based (DP-C) and Transformer-based (DP-T). Both follow the official implementation, employing DDPM with 100 denoising steps.

\begin{table}[ht]
  \centering
  \small
  \begin{tabular}{@{}lcccccc@{}}
    \toprule
    Policy & p1 & p2 & p3 & p4 & Params & Speed \\
    \midrule
    BET~\cite{shafiullah2022behavior} & 0.96 & 0.84 & 0.60 & 0.20 & \textbf{0.30} & \textbf{1.95} \\
    DP-C~\cite{chi2023diffusionpolicy} & \textbf{1.00} & \textbf{1.00} & \textbf{1.00} & 0.96 & 66.94 & 56.14 \\
    DP-T~\cite{chi2023diffusionpolicy} & \textbf{1.00} & 0.98 & 0.98 & 0.96 & 80.42 & 56.32 \\
    \rowcolor{cyan!20}
    CARP (Ours) & \textbf{1.00} & \textbf{1.00} & 0.98 & \textbf{0.98} & \textbf{3.88} & \textbf{2.01} \\
    \bottomrule
  \end{tabular}
  \vspace{-2.5mm}
  \caption{
  \textbf{Multi-Stage Task Results (State Policy).}
    In the Kitchen, p$x$ represents the success rate of interacting with $x$ or more objects.
    CARP outperforms BET, especially on challenging metrics like p4, and achieves competitive performance compared to DP, with fewer parameters and faster inference speed.}
  \label{tab:image_kitchen}
\end{table}

\begin{figure}[ht]
    \centering
    \includegraphics[width=1.0\linewidth]{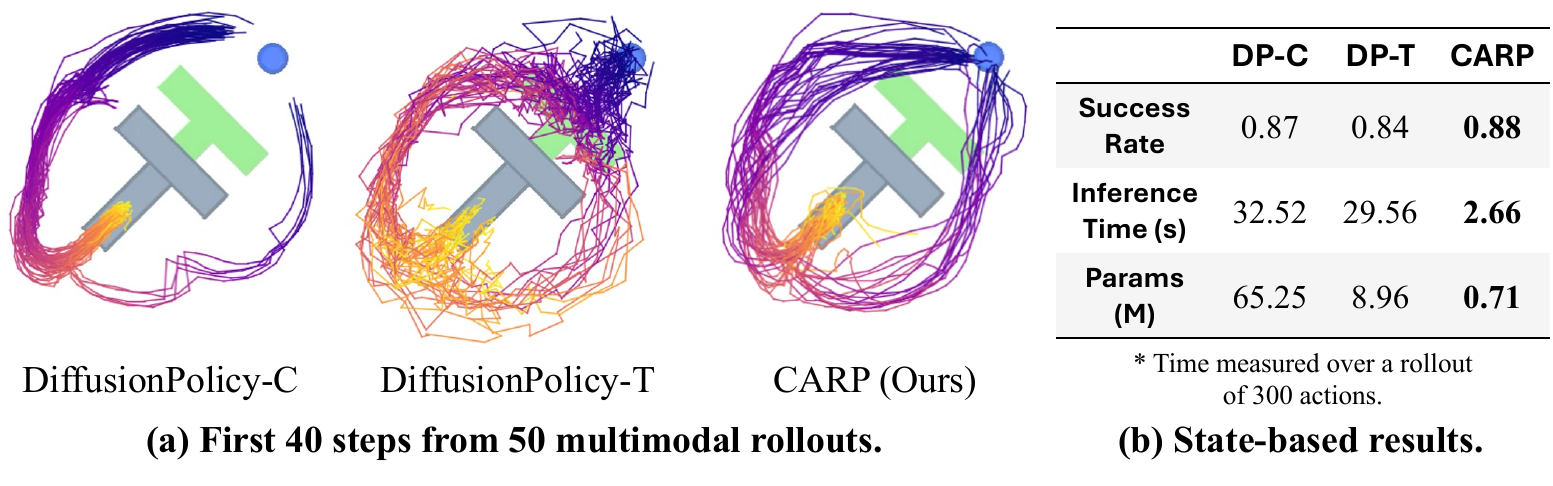}
    \vspace{-7mm}
    \caption{
    \textbf{Multi-Modal Behavior Results (State Policy).}
    On the Push-T task featuring multi-modal path options, CARP generates diverse, consistent predictions from the same initialization. Notably, it outperforms baselines in both task success and inference speed.
    }
    \label{fig:pusht_exp}
    \vspace{-2.5mm}
\end{figure}

\textbf{Metrics.} 
For each task, we evaluate policies by conducting 50 trials with random initializations, computing the success rate, and reporting the average over the top three checkpoints. 
In the Kitchen task, success rates are cumulative, requiring completion of previous levels to achieve the next (e.g., p2 requires completing p1 first).
Inference speed is measured on an A100 GPU by averaging the time taken for predicting 400 actions (280 for Kitchen, 300 for Push-T) across five runs to ensure robustness.
We also record each policy’s parameter count using the same PyTorch interface, excluding the visual encoder in the image-based setting.

\begin{figure*}[t]
    \centering
    \includegraphics[width=0.91\linewidth, keepaspectratio]{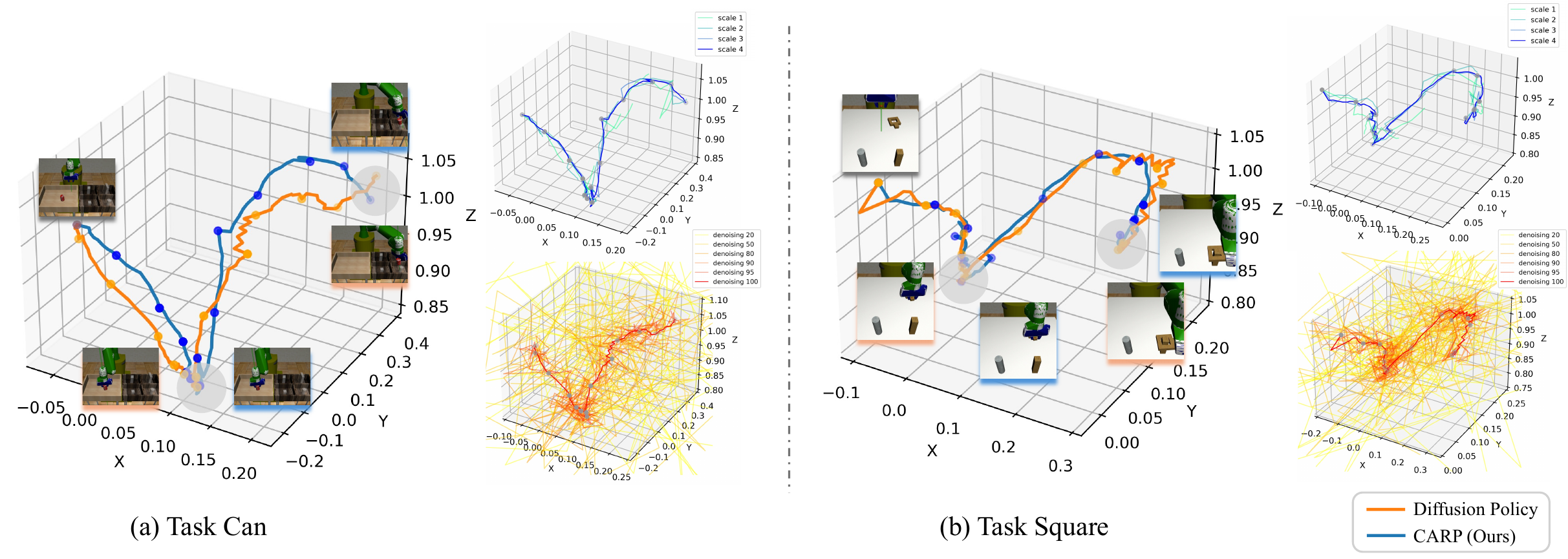}
    \vspace{-4mm}
    \caption{
    \textbf{Visualization of the Trajectory and Refining Process.}
    The left panel shows the final predicted trajectories for each task, with CARP producing smoother and more consistent paths than Diffusion Policy (DP). 
    The right panel visualizes intermediate trajectories during the refinement process for CARP (top-right) and DP (bottom-right). 
    DP displays considerable redundancy, resulting in slower processing and unstable action prediction, as illustrated by 6 selected steps among 100 denoising steps. 
    In contrast, CARP achieves efficient trajectory refinement across all 4 scales, with each step contributing meaningful updates. 
    }
    \label{fig:denoise}
\end{figure*}

\textbf{Results.}
As demonstrated in Tables \ref{tab:state_robomimic}, \ref{tab:image_robomimic}, and \ref{tab:image_kitchen}, CARP consistently outperforms autoregressive policies and demonstrates comparable performance to diffusion policies across both state-based and image-based tasks, answering \textbf{RQ1}. 
Notably, CARP significantly outperforms diffusion policies in terms of inference speed, being approximately 10 times faster, with only 1-5\% of the parameters required by diffusion models, thereby strongly supporting \textbf{RQ2} regarding CARP's efficiency. See Suppl. G for extra failure analysis. 
As shown in~\cref{fig:pusht_exp}, CARP’s use of Cross-Entropy loss at each scale during training facilitates latent-space sampling at inference, enhancing multimodality and producing smoother actions. 
This leads to superior performance and faster inference, further supporting \textbf{RQ1} and \textbf{RQ2}.

\textbf{Analysis.}
To further examine CARP's stability and efficiency, we visualize spatial trajectories along the $xyz$ axes for the Can and Square tasks in \cref{fig:denoise}. 
In each task's left panel, CARP consistently reaches specific regions (light grey area) to perform task-related actions, such as positioning for object grasping or placement. Compared to diffusion-based policies, CARP’s trajectories are smoother and more consistent, underscoring its stability and accuracy (supporting \textbf{RQ1}).
The right panels of \cref{fig:denoise} compare CARP's \textit{coarse-to-fine} action predictions with 6 selected denoising steps of the diffusion model. CARP achieves accurate predictions within just 4 \textit{coarse-to-fine} steps, whereas the diffusion model requires numerous denoising iterations, with many early steps introducing redundant computations. 
This analysis further supports \textbf{RQ2} and highlights CARP's efficiency and fast inference convergence advantage over diffusion policies in generating accurate actions with fewer refinement steps, as detailed in \cref{sec:method_ar}.
See Suppl. D, E, and F for more analysis.

\subsection{Evaluation on Multi-Task Simulations}
\label{sec:exp_sim_multi_task}

We further evaluate CARP on the MimicGen~\cite{mandlekar2023mimicgen} multi-task simulation benchmark, widely used by the state-of-the-art Sparse Diffusion Policy (SDP)~\cite{wang2024sparse}, to demonstrate CARP’s flexibility, a result of its GPT-style autoregressive design.

\begin{table*}[ht]
  \centering
  \small
  \begin{tabular}{@{}lccccccccccc@{}}
    \toprule
    Policy & Prams/M & Speed/s & Coffee & Hammer & Mug & Nut & Square & Stack & Stack three & Threading & Avg. \\
    \midrule
    TCD~\cite{liang2024skilldiffuser} & 156.11 & 107.15 & 0.77 & 0.92 & 0.53 & 0.44 & 0.63 & 0.95 & 0.62 & 0.56 & 0.68 \\
    SDP~\cite{wang2024sparse} & 159.85 & 112.39 & 0.82 & \textbf{1.00} & 0.62 & 0.54 & 0.82 & 0.96 & 0.80 & \textbf{0.70} & 0.78 \\
    \rowcolor{cyan!20}
    CARP (Ours) & \textbf{16.08} & \textbf{6.92} & \textbf{0.86} & 0.98 & \textbf{0.74} & \textbf{0.78} & \textbf{0.90} & \textbf{1.00} & \textbf{0.82} & \textbf{0.70} & \textbf{0.85} \\
    \bottomrule
  \end{tabular}
  \vspace{-2.5mm}
  \caption{
    \textbf{Multi-Task Simulation Results (Visual Policy).}
    Success rates are averaged across the top three checkpoints for each task, as well as the overall average across all tasks. We also report parameter count and inference time for generating 400 actions. CARP outperforms diffusion-based policies by 9\%-25\% in average performance, with significantly fewer parameters and over 10× faster inference.
  }
  \label{tab:sdp_multi_task}
  \vspace{-2mm}
\end{table*}

\textbf{Experimental Setup.}
MimicGen extends benchmark Robomimic~\cite{robomimic2021} by including 1K–10K human demonstrations per task, with diverse initial state distributions for enhanced generalization in multi-task evaluation. 
We select 8 robosuite~\cite{robosuite2020} tasks for evaluation, each with 1K training trajectories, following the settings in SDP~\cite{wang2024sparse}, as shown in \cref{fig:sim_task_vis}. For CARP, we simply extend the single-task implementation by incorporating a task embedding as an additional condition alongside the observation sequence $\boldsymbol{s}$. 

\textbf{Baselines.}
We compare CARP with two baselines: task-conditioned diffusion (TCD)~\cite{ajay2022conditional, liang2024skilldiffuser}: A basic diffusion-based multi-task policy and sparse diffusion policy (SDP)~\cite{wang2024sparse}: A transformer-based diffusion policy that leverages mixture of experts (MoE)~\cite{shazeer2017outrageously}. 
Both baselines are trained with visual inputs following their official implementation and settings.  

\textbf{Metrics.}
Success rates are reported for each task as the average of the best three checkpoints. 
For Nut Assembly, partial success (e.g., placing one block inside the cylinder) is assigned a score of 0.5, while full success is scored as 1. 
For all other tasks, a score of 1 is awarded only when strict success criteria are fully satisfied. 
Additionally, we calculate the average success rate across all tasks. 
To evaluate model efficiency, we report both parameter counts (excluding the identically configured visual encoder) and the inference time required to predict 400 actions on a single A100 GPU.

\textbf{Results.}
As shown in \cref{tab:sdp_multi_task}, CARP achieves up to a 25\% average improvement in success rates compared to state-of-the-art diffusion-based policies, highlighting its strong performance. Additionally, as shown in \cref{tab:sdp_multi_task}, CARP achieves over 10× faster inference speed and uses only 10\% of the parameters compared to SDP.
Leveraging its GPT-style autoregressive design~\cite{brown2020language, raffel2020exploring}, CARP can seamlessly transition from single-task to multi-task settings with minimal structural modifications~\cite{wang2024sparse}, demonstrating the flexibility of this architecture, which strongly supports \textbf{RQ3}. 
Considering fine-grained manipulation, such as Nut and Threading, CARP outperforms diffusion methods by up to 20\%, demonstrating its strong applicability in these tasks (see Suppl. L for further analysis).
These results strongly demonstrate CARP's flexibility, solidifying its role as a high-performance and high-efficiency approach for visuomotor robotic policies.

\subsection{Evaluation on Real-World}
\label{sec:exp_real}

In this section, we evaluate our approach, CARP, on real-world tasks under compute-constrained conditions, comparing its performance and efficiency against baseline methods. 

\begin{figure}[ht]
   \centering
   \includegraphics[width=1.0\linewidth]{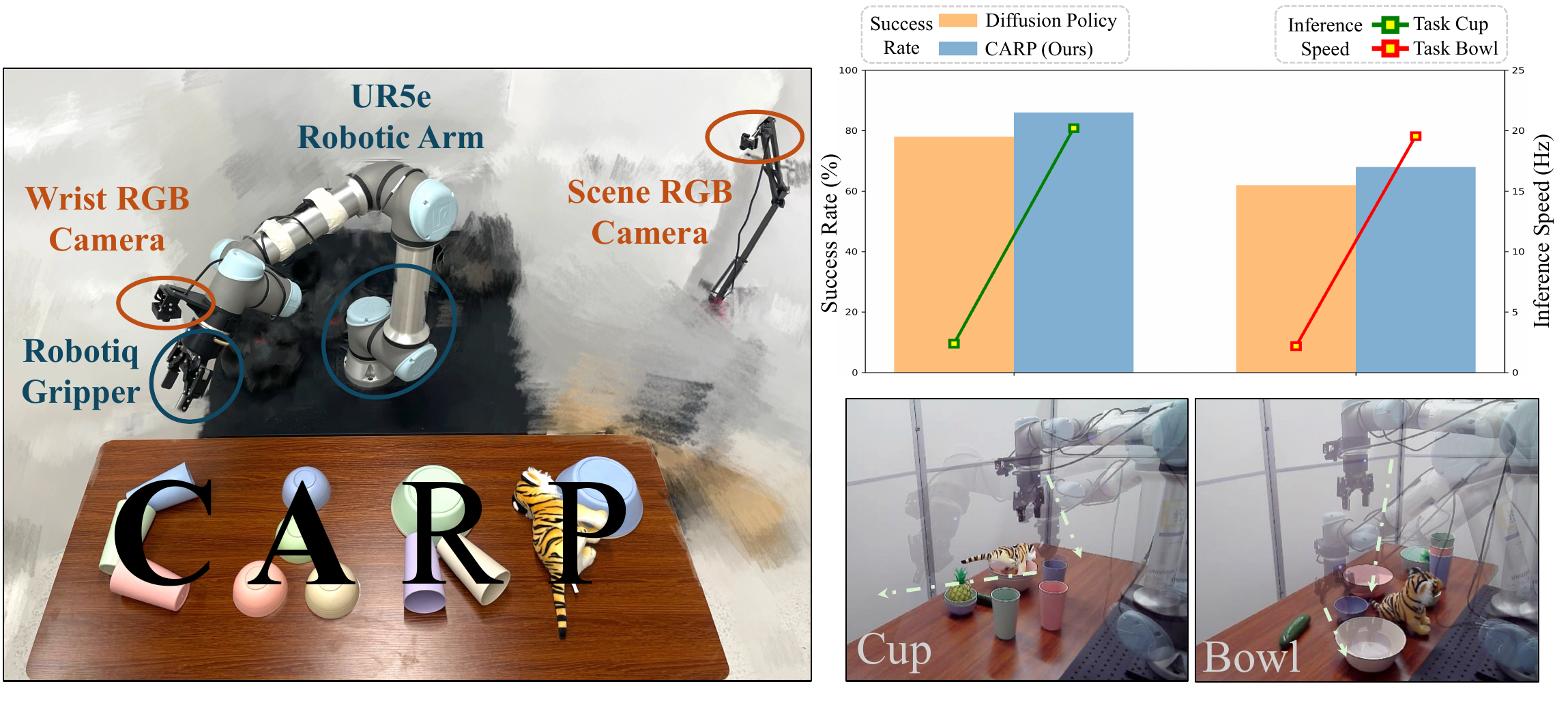}
   \vspace{-7mm}
   \caption{\textbf{Real-World Study.}
    The left panel illustrates the environment used for both experimentation and demonstration collection. The bottom-right panel visualizes trajectories from the Cup and Bowl datasets. In the top-right panel, we present the average success rate over 20 trials alongside inference speed, measured as action prediction frequency. CARP achieves competitive success rates while significantly outperforming DP in inference speed.
   }
   \label{fig:realworld_task_vis}
   \vspace{-1mm}
\end{figure}

\textbf{Experimental Setup.}
To validate CARP's real-world applicability, we design two manipulation tasks: 
1) \textit{Cup}: The robot must locate a cup on the table, pick it up, move to the right area of the table, and put it down steadily. 
2) \textit{Bowl}: The robot needs to identify a smaller bowl and a larger pot on the table, pick up the bowl, and place it inside the pot. 
We use a UR5e robotic arm with a Robotiq-2f-85 gripper, equipped with two RGB cameras: one mounted on the wrist and one in a third-person perspective (left panel, \cref{fig:realworld_task_vis}). 
The robot is controlled through 6D end-effector positioning, with inverse kinematics for joint angle calculation. 
For teleoperation, we collected 50 human demonstration trajectories for each task using a 3D Connexion space mouse. 
We reproduce the image-based CNN-based Diffusion Policy
as a baseline, as it has been shown to outperform current autoregressive policies, adapting the model's input size to match our observational setup. 
CARP maintains the design with $K=4$, consistent with previous experiments (Suppl. J for ablation studies).

\textbf{Metrics.}
For each trained policy, we report the average success rate across 20 trials per task, with the initial positions randomized. 
We also measure inference speed on an NVIDIA GeForce RTX 2060 GPU, reporting action prediction frequency in Hertz (Suppl. M for more visualization).

\textbf{Results.}
As shown in the top-right table of \cref{fig:realworld_task_vis}, CARP achieves comparable or superior performance, with up to a 10\% improvement in success rate over the Diffusion Policy across all real-world tasks, supporting \textbf{RQ1}. 
Additionally, CARP achieves approximately $8\times$ faster inference than the baseline on limited computational resources, demonstrating its suitability for real-time robotic applications, thus supporting \textbf{RQ2} (Suppl. K for additional real-world experiments).

\section{Related Work}
\label{sec:relatedwork}
\noindent \textbf{Visual Generation.}
Advances in visual generative models have strongly influenced the robotics community.
Autoregressive models generate images 
using discrete tokens from image tokenizers~\cite{huang2023towards, esser2021taming, van2017neural}.
GPT-2-style transformers~\cite{lee2022autoregressive, alexey2020image, razavi2019generating, yu2021vector} demonstrate strong performance by generating tokens sequentially.
Recent work has scaled these models to achieve impressive text-to-image synthesis results~\cite{zhao2024cobra, liu2025pite} and robotic action generation~\cite{ding2025quar, song2024germ}.
VAR~\cite{tian2024visual} introduces a new next-scale autoregressive paradigm that shifts image representation from patches to scales.
This framework~\cite{tian2024visual}, has been applied across tasks~\cite{li2024controlvar, li2024imagefolder, zhang2024var, qiu2024efficient, zhang2024g3pt, ma2024star, gu2024dart}. 
Studies~\cite{tian2024visual, sun2024autoregressive} show that autoregressive models can surpass diffusion models in achieving compatible performance, 
serving as a key inspiration for the design of our approach.

\noindent \textbf{Visuomotor Policy Learning.}
Behavior cloning~\cite{Ravichandar2020RecentAI} provides an effective approach, especially in autonomous driving~\cite{pomerleau1988alvinn} and manipulation~\cite{zhang2018deep}, offering a simpler alternative to complex reinforcement learning.
Explicit policies map observations to actions efficiently~\cite{zhang2018deep}, but struggle with complex tasks. Solutions like action discretization~\cite{zeng2021transporter} and mixture density networks~\cite{shafiullah2022behavior, cui2022play} mitigate this, yet suffer from action space explosion and hyperparameter sensitivity.
Implicit policies, such as energy-based models~\cite{jarrett2020strictly, pmlr-v164-florence22a}, offer greater flexibility but are harder to train due to optimization challenges.
Diffusion models have proven effective for policy learning~\cite{chi2023diffusionpolicy, reuss2023goal, ajay2022conditional}, but suffer from high computational cost due to multi-step denoising. Recent work aims to improve generalization~\cite{yang2024equibot}, support 3D environments~\cite{wang2024sparse, ze20243d}, and enhance modularity via mixture of experts~\cite{wang2024sparse}.
Consistency models~\cite{song2023consistency, lu2024manicm} accelerate inference but compromise action prediction accuracy, along with inflexible model design.

\section{Conclusion}
\label{sec:conclusion}

In this work, we introduce Coarse-to-Fine Autoregressive Policy (CARP), a novel paradigm for robotic visuomotor policy learning that combines the efficiency of autoregressive modeling (AM) with the high performance of diffusion modeling (DM). CARP incorporates: 
1) \textit{multi-scale} action tokenization to capture global structure and temporal locality, addressing AM's limitations in long-term dependency; 
2) \textit{coarse-to-fine} autoregressive prediction that refines actions from high-level intentions to detailed execution, achieving DM-like performance with AM-level efficiency through latent space prediction and relaxed Markovian constraints. 
The comprehensive evaluations from single- to multi-task, simulation to real-world, demonstrate CARP’s effectiveness in balancing high performance, efficiency, and flexibility.

We hope this work inspires further exploration of next-generation GPT-style autoregressive models for policy learning, fostering a unified perspective on current generative modeling techniques (see Suppl. A for further discussion).

\clearpage
\section*{Acknowledgement}

This work was supported by the National Science and Technology Innovation 2030 - Major Project (Grant No. 2022ZD0208800)

{\small
\bibliographystyle{ieeenat_fullname}
\bibliography{sec/11_references}

\begin{thebibliography}{74}
\providecommand{\natexlab}[1]{#1}
\providecommand{\url}[1]{\texttt{#1}}
\expandafter\ifx\csname urlstyle\endcsname\relax
  \providecommand{\doi}[1]{doi: #1}\else
  \providecommand{\doi}{doi: \begingroup \urlstyle{rm}\Url}\fi

\bibitem[Ahn et~al.(2022)Ahn, Brohan, Brown, Chebotar, Cortes, David, Finn, Fu, Gopalakrishnan, Hausman, et~al.]{ahn2022can}
Michael Ahn, Anthony Brohan, Noah Brown, Yevgen Chebotar, Omar Cortes, Byron David, Chelsea Finn, Chuyuan Fu, Keerthana Gopalakrishnan, Karol Hausman, et~al.
\newblock Do as i can, not as i say: Grounding language in robotic affordances.
\newblock \emph{arXiv preprint arXiv:2204.01691}, 2022.

\bibitem[Ajay et~al.(2022)Ajay, Du, Gupta, Tenenbaum, Jaakkola, and Agrawal]{ajay2022conditional}
Anurag Ajay, Yilun Du, Abhi Gupta, Joshua Tenenbaum, Tommi Jaakkola, and Pulkit Agrawal.
\newblock Is conditional generative modeling all you need for decision-making?
\newblock \emph{arXiv preprint arXiv:2211.15657}, 2022.

\bibitem[Alayrac et~al.(2022)Alayrac, Donahue, Luc, Miech, Barr, Hasson, Lenc, Mensch, Millican, Reynolds, et~al.]{alayrac2022flamingo}
Jean-Baptiste Alayrac, Jeff Donahue, Pauline Luc, Antoine Miech, Iain Barr, Yana Hasson, Karel Lenc, Arthur Mensch, Katherine Millican, Malcolm Reynolds, et~al.
\newblock Flamingo: a visual language model for few-shot learning.
\newblock \emph{Advances in neural information processing systems}, 35:\penalty0 23716--23736, 2022.

\bibitem[Alexey(2020)]{alexey2020image}
Dosovitskiy Alexey.
\newblock An image is worth 16x16 words: Transformers for image recognition at scale.
\newblock \emph{arXiv preprint arXiv: 2010.11929}, 2020.

\bibitem[Ankile et~al.(2024)Ankile, Simeonov, Shenfeld, Torne, and Agrawal]{ankile2024imitation}
Lars Ankile, Anthony Simeonov, Idan Shenfeld, Marcel Torne, and Pulkit Agrawal.
\newblock From imitation to refinement--residual rl for precise assembly.
\newblock \emph{arXiv preprint arXiv:2407.16677}, 2024.

\bibitem[Bojarski(2016)]{bojarski2016end}
Mariusz Bojarski.
\newblock End to end learning for self-driving cars.
\newblock \emph{arXiv preprint arXiv:1604.07316}, 2016.

\bibitem[Brohan et~al.(2023)Brohan, Brown, Carbajal, Chebotar, Chen, Choromanski, Ding, Driess, Dubey, Finn, et~al.]{brohan2023rt}
Anthony Brohan, Noah Brown, Justice Carbajal, Yevgen Chebotar, Xi Chen, Krzysztof Choromanski, Tianli Ding, Danny Driess, Avinava Dubey, Chelsea Finn, et~al.
\newblock Rt-2: Vision-language-action models transfer web knowledge to robotic control.
\newblock \emph{arXiv preprint arXiv:2307.15818}, 2023.

\bibitem[Brown(2020)]{brown2020language}
Tom~B Brown.
\newblock Language models are few-shot learners.
\newblock \emph{arXiv preprint arXiv:2005.14165}, 2020.

\bibitem[chaandar Ravichandar et~al.(2020)chaandar Ravichandar, Polydoros, Chernova, and Billard]{Ravichandar2020RecentAI}
Harish chaandar Ravichandar, Athanasios~S. Polydoros, Sonia Chernova, and Aude Billard.
\newblock Recent advances in robot learning from demonstration.
\newblock \emph{Annu. Rev. Control. Robotics Auton. Syst.}, 3:\penalty0 297--330, 2020.

\bibitem[Chen et~al.(2023)Chen, Yu, Ge, Yao, Xie, Wu, Wang, Kwok, Luo, Lu, et~al.]{chen2023pixart}
Junsong Chen, Jincheng Yu, Chongjian Ge, Lewei Yao, Enze Xie, Yue Wu, Zhongdao Wang, James Kwok, Ping Luo, Huchuan Lu, et~al.
\newblock Pixart-alpha: Fast training of diffusion transformer for photorealistic text-to-image synthesis.
\newblock \emph{arXiv preprint arXiv:2310.00426}, 2023.

\bibitem[Chen et~al.(2021)Chen, Lu, Rajeswaran, Lee, Grover, Laskin, Abbeel, Srinivas, and Mordatch]{chen2021decision}
Lili Chen, Kevin Lu, Aravind Rajeswaran, Kimin Lee, Aditya Grover, Misha Laskin, Pieter Abbeel, Aravind Srinivas, and Igor Mordatch.
\newblock Decision transformer: Reinforcement learning via sequence modeling.
\newblock \emph{Advances in neural information processing systems}, 34:\penalty0 15084--15097, 2021.

\bibitem[Chi et~al.(2023)Chi, Feng, Du, Xu, Cousineau, Burchfiel, and Song]{chi2023diffusionpolicy}
Cheng Chi, Siyuan Feng, Yilun Du, Zhenjia Xu, Eric Cousineau, Benjamin Burchfiel, and Shuran Song.
\newblock Diffusion policy: Visuomotor policy learning via action diffusion.
\newblock In \emph{Proceedings of Robotics: Science and Systems (RSS)}, 2023.

\bibitem[Cui et~al.(2022)Cui, Wang, Shafiullah, and Pinto]{cui2022play}
Zichen~Jeff Cui, Yibin Wang, Nur Muhammad~Mahi Shafiullah, and Lerrel Pinto.
\newblock From play to policy: Conditional behavior generation from uncurated robot data.
\newblock \emph{arXiv preprint arXiv:2210.10047}, 2022.

\bibitem[Ding et~al.(2025)Ding, Zhao, Zhang, Song, Zhang, Huang, Yang, and Wang]{ding2025quar}
Pengxiang Ding, Han Zhao, Wenjie Zhang, Wenxuan Song, Min Zhang, Siteng Huang, Ningxi Yang, and Donglin Wang.
\newblock Quar-vla: Vision-language-action model for quadruped robots.
\newblock In \emph{European Conference on Computer Vision}, pages 352--367. Springer, 2025.

\bibitem[Esser et~al.(2021)Esser, Rombach, and Ommer]{esser2021taming}
Patrick Esser, Robin Rombach, and Bjorn Ommer.
\newblock Taming transformers for high-resolution image synthesis.
\newblock In \emph{Proceedings of the IEEE/CVF conference on computer vision and pattern recognition}, pages 12873--12883, 2021.

\bibitem[Florence et~al.(2022)Florence, Lynch, Zeng, Ramirez, Wahid, Downs, Wong, Lee, Mordatch, and Tompson]{pmlr-v164-florence22a}
Pete Florence, Corey Lynch, Andy Zeng, Oscar~A Ramirez, Ayzaan Wahid, Laura Downs, Adrian Wong, Johnny Lee, Igor Mordatch, and Jonathan Tompson.
\newblock Implicit behavioral cloning.
\newblock In \emph{Proceedings of the 5th Conference on Robot Learning}, pages 158--168. PMLR, 2022.

\bibitem[Gu et~al.(2024)Gu, Wang, Zhang, Zhang, Zhang, Jaitly, Susskind, and Zhai]{gu2024dart}
Jiatao Gu, Yuyang Wang, Yizhe Zhang, Qihang Zhang, Dinghuai Zhang, Navdeep Jaitly, Josh Susskind, and Shuangfei Zhai.
\newblock Dart: Denoising autoregressive transformer for scalable text-to-image generation.
\newblock \emph{arXiv preprint arXiv:2410.08159}, 2024.

\bibitem[Gupta et~al.(2019)Gupta, Kumar, Lynch, Levine, and Hausman]{gupta2019relay}
Abhishek Gupta, Vikash Kumar, Corey Lynch, Sergey Levine, and Karol Hausman.
\newblock Relay policy learning: Solving long-horizon tasks via imitation and reinforcement learning.
\newblock \emph{arXiv preprint arXiv:1910.11956}, 2019.

\bibitem[Haynes et~al.(2012)Haynes, Corns, and Venayagamoorthy]{haynes2012exponential}
David Haynes, Steven Corns, and Ganesh~Kumar Venayagamoorthy.
\newblock An exponential moving average algorithm.
\newblock In \emph{2012 IEEE Congress on Evolutionary Computation}, pages 1--8. IEEE, 2012.

\bibitem[Heo et~al.(2023)Heo, Lee, Lee, and Lim]{heo2023furniturebench}
Minho Heo, Youngwoon Lee, Doohyun Lee, and Joseph~J Lim.
\newblock Furniturebench: Reproducible real-world benchmark for long-horizon complex manipulation.
\newblock \emph{The International Journal of Robotics Research}, page 02783649241304789, 2023.

\bibitem[Ho et~al.(2020)Ho, Jain, and Abbeel]{ho2020denoising}
Jonathan Ho, Ajay Jain, and Pieter Abbeel.
\newblock Denoising diffusion probabilistic models.
\newblock \emph{arXiv preprint arxiv:2006.11239}, 2020.

\bibitem[Huang et~al.(2023)Huang, Mao, Chen, and Zhang]{huang2023towards}
Mengqi Huang, Zhendong Mao, Zhuowei Chen, and Yongdong Zhang.
\newblock Towards accurate image coding: Improved autoregressive image generation with dynamic vector quantization.
\newblock In \emph{Proceedings of the IEEE/CVF Conference on Computer Vision and Pattern Recognition}, pages 22596--22605, 2023.

\bibitem[Janner et~al.(2021)Janner, Li, and Levine]{janner2021offline}
Michael Janner, Qiyang Li, and Sergey Levine.
\newblock Offline reinforcement learning as one big sequence modeling problem.
\newblock \emph{Advances in neural information processing systems}, 34:\penalty0 1273--1286, 2021.

\bibitem[Janner et~al.(2022)Janner, Du, Tenenbaum, and Levine]{janner2022diffuser}
Michael Janner, Yilun Du, Joshua Tenenbaum, and Sergey Levine.
\newblock Planning with diffusion for flexible behavior synthesis.
\newblock In \emph{International Conference on Machine Learning}, 2022.

\bibitem[Jarrett et~al.(2020)Jarrett, Bica, and van~der Schaar]{jarrett2020strictly}
Daniel Jarrett, Ioana Bica, and Mihaela van~der Schaar.
\newblock Strictly batch imitation learning by energy-based distribution matching.
\newblock \emph{Advances in Neural Information Processing Systems}, 33:\penalty0 7354--7365, 2020.

\bibitem[Jiang et~al.(2022)Jiang, Gupta, Zhang, Wang, Dou, Chen, Fei-Fei, Anandkumar, Zhu, and Fan]{jiang2022vima}
Yunfan Jiang, Agrim Gupta, Zichen Zhang, Guanzhi Wang, Yongqiang Dou, Yanjun Chen, Li Fei-Fei, Anima Anandkumar, Yuke Zhu, and Linxi Fan.
\newblock Vima: General robot manipulation with multimodal prompts.
\newblock \emph{arXiv preprint arXiv:2210.03094}, 2\penalty0 (3):\penalty0 6, 2022.

\bibitem[Kaplan et~al.(2020)Kaplan, McCandlish, Henighan, Brown, Chess, Child, Gray, Radford, Wu, and Amodei]{kaplan2020scaling}
Jared Kaplan, Sam McCandlish, Tom Henighan, Tom~B Brown, Benjamin Chess, Rewon Child, Scott Gray, Alec Radford, Jeffrey Wu, and Dario Amodei.
\newblock Scaling laws for neural language models.
\newblock \emph{arXiv preprint arXiv:2001.08361}, 2020.

\bibitem[Kilian et~al.(2024)Kilian, Jampani, and Zettlemoyer]{kilian2024computational}
Maciej Kilian, Varun Jampani, and Luke Zettlemoyer.
\newblock Computational tradeoffs in image synthesis: Diffusion, masked-token, and next-token prediction.
\newblock \emph{arXiv preprint arXiv:2405.13218}, 2024.

\bibitem[Kim et~al.(2024)Kim, Pertsch, Karamcheti, Xiao, Balakrishna, Nair, Rafailov, Foster, Lam, Sanketi, et~al.]{kim2024openvla}
Moo~Jin Kim, Karl Pertsch, Siddharth Karamcheti, Ted Xiao, Ashwin Balakrishna, Suraj Nair, Rafael Rafailov, Ethan Foster, Grace Lam, Pannag Sanketi, et~al.
\newblock Openvla: An open-source vision-language-action model.
\newblock \emph{arXiv preprint arXiv:2406.09246}, 2024.

\bibitem[Lai et~al.(2022)Lai, Huang, and Gershman]{lai2022action}
Lucy Lai, Ann~Zixiang Huang, and Samuel~J Gershman.
\newblock Action chunking as policy compression.
\newblock \emph{PsyArXiv}, 2022.

\bibitem[Lee et~al.(2022)Lee, Kim, Kim, Cho, and Han]{lee2022autoregressive}
Doyup Lee, Chiheon Kim, Saehoon Kim, Minsu Cho, and Wook-Shin Han.
\newblock Autoregressive image generation using residual quantization.
\newblock In \emph{Proceedings of the IEEE/CVF Conference on Computer Vision and Pattern Recognition}, pages 11523--11532, 2022.

\bibitem[Lee et~al.(2024)Lee, Wang, Etukuru, Kim, Shafiullah, and Pinto]{lee2024behavior}
Seungjae Lee, Yibin Wang, Haritheja Etukuru, H~Jin Kim, Nur Muhammad~Mahi Shafiullah, and Lerrel Pinto.
\newblock Behavior generation with latent actions.
\newblock \emph{arXiv preprint arXiv:2403.03181}, 2024.

\bibitem[Li et~al.(2023)Li, Liu, Zhang, Yu, Xu, Wu, Cheang, Jing, Zhang, Liu, et~al.]{li2023vision}
Xinghang Li, Minghuan Liu, Hanbo Zhang, Cunjun Yu, Jie Xu, Hongtao Wu, Chilam Cheang, Ya Jing, Weinan Zhang, Huaping Liu, et~al.
\newblock Vision-language foundation models as effective robot imitators.
\newblock \emph{arXiv preprint arXiv:2311.01378}, 2023.

\bibitem[Li et~al.(2024{\natexlab{a}})Li, Chen, Qiu, Kuen, Gu, Raj, and Lin]{li2024imagefolder}
Xiang Li, Hao Chen, Kai Qiu, Jason Kuen, Jiuxiang Gu, Bhiksha Raj, and Zhe Lin.
\newblock Imagefolder: Autoregressive image generation with folded tokens.
\newblock \emph{arXiv preprint arXiv:2410.01756}, 2024{\natexlab{a}}.

\bibitem[Li et~al.(2024{\natexlab{b}})Li, Qiu, Chen, Kuen, Lin, Singh, and Raj]{li2024controlvar}
Xiang Li, Kai Qiu, Hao Chen, Jason Kuen, Zhe Lin, Rita Singh, and Bhiksha Raj.
\newblock Controlvar: Exploring controllable visual autoregressive modeling.
\newblock \emph{arXiv preprint arXiv:2406.09750}, 2024{\natexlab{b}}.

\bibitem[Liang et~al.(2024)Liang, Mu, Ma, Tomizuka, Ding, and Luo]{liang2024skilldiffuser}
Zhixuan Liang, Yao Mu, Hengbo Ma, Masayoshi Tomizuka, Mingyu Ding, and Ping Luo.
\newblock Skilldiffuser: Interpretable hierarchical planning via skill abstractions in diffusion-based task execution.
\newblock In \emph{Proceedings of the IEEE/CVF Conference on Computer Vision and Pattern Recognition}, pages 16467--16476, 2024.

\bibitem[Liu et~al.(2025)Liu, Ding, Huang, Zhang, Zhao, and Wang]{liu2025pite}
Yang Liu, Pengxiang Ding, Siteng Huang, Min Zhang, Han Zhao, and Donglin Wang.
\newblock Pite: Pixel-temporal alignment for large video-language model.
\newblock In \emph{European Conference on Computer Vision}, pages 160--176. Springer, 2025.

\bibitem[Lu et~al.(2024)Lu, Gao, Chen, Dai, Wang, and Tang]{lu2024manicm}
Guanxing Lu, Zifeng Gao, Tianxing Chen, Wenxun Dai, Ziwei Wang, and Yansong Tang.
\newblock Manicm: Real-time 3d diffusion policy via consistency model for robotic manipulation.
\newblock \emph{arXiv preprint arXiv:2406.01586}, 2024.

\bibitem[Ma et~al.(2024)Ma, Zhou, Liang, Bai, Zhao, Chen, and Jin]{ma2024star}
Xiaoxiao Ma, Mohan Zhou, Tao Liang, Yalong Bai, Tiejun Zhao, Huaian Chen, and Yi Jin.
\newblock Star: Scale-wise text-to-image generation via auto-regressive representations.
\newblock \emph{arXiv preprint arXiv:2406.10797}, 2024.

\bibitem[Mandlekar et~al.(2021)Mandlekar, Xu, Wong, Nasiriany, Wang, Kulkarni, Fei-Fei, Savarese, Zhu, and Mart\'{i}n-Mart\'{i}n]{robomimic2021}
Ajay Mandlekar, Danfei Xu, Josiah Wong, Soroush Nasiriany, Chen Wang, Rohun Kulkarni, Li Fei-Fei, Silvio Savarese, Yuke Zhu, and Roberto Mart\'{i}n-Mart\'{i}n.
\newblock What matters in learning from offline human demonstrations for robot manipulation.
\newblock In \emph{arXiv preprint arXiv:2108.03298}, 2021.

\bibitem[Mandlekar et~al.(2023)Mandlekar, Nasiriany, Wen, Akinola, Narang, Fan, Zhu, and Fox]{mandlekar2023mimicgen}
Ajay Mandlekar, Soroush Nasiriany, Bowen Wen, Iretiayo Akinola, Yashraj Narang, Linxi Fan, Yuke Zhu, and Dieter Fox.
\newblock Mimicgen: A data generation system for scalable robot learning using human demonstrations.
\newblock In \emph{7th Annual Conference on Robot Learning}, 2023.

\bibitem[Park et~al.(2019)Park, Liu, Wang, and Zhu]{park2019semantic}
Taesung Park, Ming-Yu Liu, Ting-Chun Wang, and Jun-Yan Zhu.
\newblock Semantic image synthesis with spatially-adaptive normalization.
\newblock In \emph{Proceedings of the IEEE/CVF conference on computer vision and pattern recognition}, pages 2337--2346, 2019.

\bibitem[Pomerleau(1988)]{pomerleau1988alvinn}
Dean~A Pomerleau.
\newblock Alvinn: An autonomous land vehicle in a neural network.
\newblock \emph{Advances in neural information processing systems}, 1, 1988.

\bibitem[Prasad et~al.(2024)Prasad, Lin, Wu, Zhou, and Bohg]{prasad2024consistency}
Aaditya Prasad, Kevin Lin, Jimmy Wu, Linqi Zhou, and Jeannette Bohg.
\newblock Consistency policy: Accelerated visuomotor policies via consistency distillation.
\newblock \emph{arXiv preprint arXiv:2405.07503}, 2024.

\bibitem[Qiu et~al.(2024)Qiu, Li, Chen, Sun, Wang, Lin, Savvides, and Raj]{qiu2024efficient}
Kai Qiu, Xiang Li, Hao Chen, Jie Sun, Jinglu Wang, Zhe Lin, Marios Savvides, and Bhiksha Raj.
\newblock Efficient autoregressive audio modeling via next-scale prediction.
\newblock \emph{arXiv preprint arXiv:2408.09027}, 2024.

\bibitem[Radford et~al.(2019)Radford, Wu, Child, Luan, Amodei, Sutskever, et~al.]{radford2019language}
Alec Radford, Jeffrey Wu, Rewon Child, David Luan, Dario Amodei, Ilya Sutskever, et~al.
\newblock Language models are unsupervised multitask learners.
\newblock \emph{OpenAI blog}, 1\penalty0 (8):\penalty0 9, 2019.

\bibitem[Radford et~al.(2021)Radford, Kim, Hallacy, Ramesh, Goh, Agarwal, Sastry, Askell, Mishkin, Clark, et~al.]{radford2021learning}
Alec Radford, Jong~Wook Kim, Chris Hallacy, Aditya Ramesh, Gabriel Goh, Sandhini Agarwal, Girish Sastry, Amanda Askell, Pamela Mishkin, Jack Clark, et~al.
\newblock Learning transferable visual models from natural language supervision.
\newblock In \emph{International conference on machine learning}, pages 8748--8763. PMLR, 2021.

\bibitem[Radosavovic et~al.(2023)Radosavovic, Shi, Fu, Goldberg, Darrell, and Malik]{Rpt2023}
Ilija Radosavovic, Baifeng Shi, Letian Fu, Ken Goldberg, Trevor Darrell, and Jitendra Malik.
\newblock Robot learning with sensorimotor pre-training.
\newblock \emph{arXiv:2306.10007}, 2023.

\bibitem[Raffel et~al.(2020)Raffel, Shazeer, Roberts, Lee, Narang, Matena, Zhou, Li, and Liu]{raffel2020exploring}
Colin Raffel, Noam Shazeer, Adam Roberts, Katherine Lee, Sharan Narang, Michael Matena, Yanqi Zhou, Wei Li, and Peter~J Liu.
\newblock Exploring the limits of transfer learning with a unified text-to-text transformer.
\newblock \emph{Journal of machine learning research}, 21\penalty0 (140):\penalty0 1--67, 2020.

\bibitem[Razavi et~al.(2019)Razavi, Van~den Oord, and Vinyals]{razavi2019generating}
Ali Razavi, Aaron Van~den Oord, and Oriol Vinyals.
\newblock Generating diverse high-fidelity images with vq-vae-2.
\newblock \emph{Advances in neural information processing systems}, 32, 2019.

\bibitem[Reed et~al.(2022)Reed, Zolna, Parisotto, Colmenarejo, Novikov, Barth-Maron, Gimenez, Sulsky, Kay, Springenberg, et~al.]{reed2022generalist}
Scott Reed, Konrad Zolna, Emilio Parisotto, Sergio~Gomez Colmenarejo, Alexander Novikov, Gabriel Barth-Maron, Mai Gimenez, Yury Sulsky, Jackie Kay, Jost~Tobias Springenberg, et~al.
\newblock A generalist agent.
\newblock \emph{arXiv preprint arXiv:2205.06175}, 2022.

\bibitem[Reuss et~al.(2023)Reuss, Li, Jia, and Lioutikov]{reuss2023goal}
Moritz Reuss, Maximilian Li, Xiaogang Jia, and Rudolf Lioutikov.
\newblock Goal-conditioned imitation learning using score-based diffusion policies.
\newblock \emph{arXiv preprint arXiv:2304.02532}, 2023.

\bibitem[Reuss et~al.(2024)Reuss, Ya{\u{g}}murlu, Wenzel, and Lioutikov]{reuss2024multimodal}
Moritz Reuss, {\"O}mer~Erdin{\c{c}} Ya{\u{g}}murlu, Fabian Wenzel, and Rudolf Lioutikov.
\newblock Multimodal diffusion transformer: Learning versatile behavior from multimodal goals.
\newblock In \emph{First Workshop on Vision-Language Models for Navigation and Manipulation at ICRA 2024}, 2024.

\bibitem[Shafiullah et~al.(2022)Shafiullah, Cui, Altanzaya, and Pinto]{shafiullah2022behavior}
Nur Muhammad~Mahi Shafiullah, Zichen~Jeff Cui, Ariuntuya Altanzaya, and Lerrel Pinto.
\newblock Behavior transformers: Cloning $k$ modes with one stone.
\newblock In \emph{Thirty-Sixth Conference on Neural Information Processing Systems}, 2022.

\bibitem[Shazeer et~al.(2017)Shazeer, Mirhoseini, Maziarz, Davis, Le, Hinton, and Dean]{shazeer2017outrageously}
Noam Shazeer, Azalia Mirhoseini, Krzysztof Maziarz, Andy Davis, Quoc Le, Geoffrey Hinton, and Jeff Dean.
\newblock Outrageously large neural networks: The sparsely-gated mixture-of-experts layer.
\newblock \emph{arXiv preprint arXiv:1701.06538}, 2017.

\bibitem[Song et~al.(2024)Song, Zhao, Ding, Cui, Lyu, Fan, and Wang]{song2024germ}
Wenxuan Song, Han Zhao, Pengxiang Ding, Can Cui, Shangke Lyu, Yaning Fan, and Donglin Wang.
\newblock Germ: A generalist robotic model with mixture-of-experts for quadruped robot.
\newblock \emph{arXiv preprint arXiv:2403.13358}, 2024.

\bibitem[Song et~al.(2023)Song, Dhariwal, Chen, and Sutskever]{song2023consistency}
Yang Song, Prafulla Dhariwal, Mark Chen, and Ilya Sutskever.
\newblock Consistency models.
\newblock \emph{arXiv preprint arXiv:2303.01469}, 2023.

\bibitem[Sun et~al.(2024)Sun, Jiang, Chen, Zhang, Peng, Luo, and Yuan]{sun2024autoregressive}
Peize Sun, Yi Jiang, Shoufa Chen, Shilong Zhang, Bingyue Peng, Ping Luo, and Zehuan Yuan.
\newblock Autoregressive model beats diffusion: Llama for scalable image generation.
\newblock \emph{arXiv preprint arXiv:2406.06525}, 2024.

\bibitem[Thomas et~al.(2023)Thomas, Cheng, Loynd, Frujeri, Vineet, Jalobeanu, and Kolobov]{thomas2023plex}
Garrett Thomas, Ching-An Cheng, Ricky Loynd, Felipe~Vieira Frujeri, Vibhav Vineet, Mihai Jalobeanu, and Andrey Kolobov.
\newblock Plex: Making the most of the available data for robotic manipulation pretraining.
\newblock In \emph{Conference on Robot Learning}, pages 2624--2641. PMLR, 2023.

\bibitem[Tian et~al.(2024)Tian, Jiang, Yuan, Peng, and Wang]{tian2024visual}
Keyu Tian, Yi Jiang, Zehuan Yuan, Bingyue Peng, and Liwei Wang.
\newblock Visual autoregressive modeling: Scalable image generation via next-scale prediction.
\newblock \emph{arXiv preprint arXiv:2404.02905}, 2024.

\bibitem[Torabi et~al.(2018)Torabi, Warnell, and Stone]{torabi2018behavioral}
Faraz Torabi, Garrett Warnell, and Peter Stone.
\newblock Behavioral cloning from observation.
\newblock \emph{arXiv preprint arXiv:1805.01954}, 2018.

\bibitem[Van Den~Oord et~al.(2017)Van Den~Oord, Vinyals, et~al.]{van2017neural}
Aaron Van Den~Oord, Oriol Vinyals, et~al.
\newblock Neural discrete representation learning.
\newblock \emph{Advances in neural information processing systems}, 30, 2017.

\bibitem[Wang et~al.(2024)Wang, Zhang, Huo, Tian, Zhang, Xie, Xu, Ji, Zhan, Ding, et~al.]{wang2024sparse}
Yixiao Wang, Yifei Zhang, Mingxiao Huo, Ran Tian, Xiang Zhang, Yichen Xie, Chenfeng Xu, Pengliang Ji, Wei Zhan, Mingyu Ding, et~al.
\newblock Sparse diffusion policy: A sparse, reusable, and flexible policy for robot learning.
\newblock \emph{arXiv preprint arXiv:2407.01531}, 2024.

\bibitem[Yang et~al.(2024)Yang, Cao, Deng, Antonova, Song, and Bohg]{yang2024equibot}
Jingyun Yang, Zi-ang Cao, Congyue Deng, Rika Antonova, Shuran Song, and Jeannette Bohg.
\newblock Equibot: Sim (3)-equivariant diffusion policy for generalizable and data efficient learning.
\newblock \emph{arXiv preprint arXiv:2407.01479}, 2024.

\bibitem[Yu et~al.(2021)Yu, Li, Koh, Zhang, Pang, Qin, Ku, Xu, Baldridge, and Wu]{yu2021vector}
Jiahui Yu, Xin Li, Jing~Yu Koh, Han Zhang, Ruoming Pang, James Qin, Alexander Ku, Yuanzhong Xu, Jason Baldridge, and Yonghui Wu.
\newblock Vector-quantized image modeling with improved vqgan.
\newblock \emph{arXiv preprint arXiv:2110.04627}, 2021.

\bibitem[Ze et~al.(2024)Ze, Zhang, Zhang, Hu, Wang, and Xu]{ze20243d}
Yanjie Ze, Gu Zhang, Kangning Zhang, Chenyuan Hu, Muhan Wang, and Huazhe Xu.
\newblock 3d diffusion policy: Generalizable visuomotor policy learning via simple 3d representations.
\newblock In \emph{ICRA 2024 Workshop on 3D Visual Representations for Robot Manipulation}, 2024.

\bibitem[Zeng et~al.(2021)Zeng, Florence, Tompson, Welker, Chien, Attarian, Armstrong, Krasin, Duong, Sindhwani, et~al.]{zeng2021transporter}
Andy Zeng, Pete Florence, Jonathan Tompson, Stefan Welker, Jonathan Chien, Maria Attarian, Travis Armstrong, Ivan Krasin, Dan Duong, Vikas Sindhwani, et~al.
\newblock Transporter networks: Rearranging the visual world for robotic manipulation.
\newblock In \emph{Conference on Robot Learning}, pages 726--747. PMLR, 2021.

\bibitem[Zhang et~al.(2024{\natexlab{a}})Zhang, Xiong, and Xu]{zhang2024g3pt}
Jinzhi Zhang, Feng Xiong, and Mu Xu.
\newblock G3pt: Unleash the power of autoregressive modeling in 3d generation via cross-scale querying transformer.
\newblock \emph{arXiv preprint arXiv:2409.06322}, 2024{\natexlab{a}}.

\bibitem[Zhang et~al.(2024{\natexlab{b}})Zhang, Dai, Yang, An, Feng, and Ren]{zhang2024var}
Qian Zhang, Xiangzi Dai, Ninghua Yang, Xiang An, Ziyong Feng, and Xingyu Ren.
\newblock Var-clip: Text-to-image generator with visual auto-regressive modeling.
\newblock \emph{arXiv preprint arXiv:2408.01181}, 2024{\natexlab{b}}.

\bibitem[Zhang et~al.(2018)Zhang, McCarthy, Jow, Lee, Chen, Goldberg, and Abbeel]{zhang2018deep}
Tianhao Zhang, Zoe McCarthy, Owen Jow, Dennis Lee, Xi Chen, Ken Goldberg, and Pieter Abbeel.
\newblock Deep imitation learning for complex manipulation tasks from virtual reality teleoperation.
\newblock In \emph{2018 IEEE international conference on robotics and automation (ICRA)}, pages 5628--5635. IEEE, 2018.

\bibitem[Zhao et~al.(2024)Zhao, Zhang, Zhao, Ding, Huang, and Wang]{zhao2024cobra}
Han Zhao, Min Zhang, Wei Zhao, Pengxiang Ding, Siteng Huang, and Donglin Wang.
\newblock Cobra: Extending mamba to multi-modal large language model for efficient inference.
\newblock \emph{arXiv preprint arXiv:2403.14520}, 2024.

\bibitem[Zhao et~al.(2023)Zhao, Kumar, Levine, and Finn]{zhao2023learning}
Tony~Z Zhao, Vikash Kumar, Sergey Levine, and Chelsea Finn.
\newblock Learning fine-grained bimanual manipulation with low-cost hardware.
\newblock \emph{arXiv preprint arXiv:2304.13705}, 2023.

\bibitem[Zhou et~al.(2019)Zhou, Barnes, Lu, Yang, and Li]{zhou2019continuity}
Yi Zhou, Connelly Barnes, Jingwan Lu, Jimei Yang, and Hao Li.
\newblock On the continuity of rotation representations in neural networks.
\newblock In \emph{Proceedings of the IEEE/CVF conference on computer vision and pattern recognition}, pages 5745--5753, 2019.

\bibitem[Zhu et~al.(2020)Zhu, Wong, Mandlekar, Mart\'{i}n-Mart\'{i}n, Joshi, Nasiriany, Zhu, and Lin]{robosuite2020}
Yuke Zhu, Josiah Wong, Ajay Mandlekar, Roberto Mart\'{i}n-Mart\'{i}n, Abhishek Joshi, Soroush Nasiriany, Yifeng Zhu, and Kevin Lin.
\newblock robosuite: A modular simulation framework and benchmark for robot learning.
\newblock In \emph{arXiv preprint arXiv:2009.12293}, 2020.

\end{thebibliography}
}

\ifarxiv \clearpage \appendix

\section{Limitations and Future Work}
\label{sec:suppl_limit_future}

In this work, we propose CARP, a next-generation paradigm for robotic visuomotor policy learning, which effectively balances the long-standing trade-off between high performance and high inference efficiency seen in previous autoregressive modeling (AM) and diffusion modeling (DM) approaches. Despite these advancements, there remain several limitations and opportunities for improvement in near-future research.

\textbf{First}, the architectural design of CARP can be further optimized for simplicity. Currently, CARP employs a two-stage design, where the first stage utilizes separate \textit{multi-scale} action VQVAE modules for each action dimension to address their orthogonality. 
A promising direction for future work could focus on developing a unified one-stage method that integrates \textit{multi-scale} tokenization with the \textit{coarse-to-fine} prediction process, resulting in a more efficient and streamlined framework without compromising performance.

\textbf{Second}, CARP's multimodal capacity has not yet been fully explored or leveraged. To address the inherent unimodality of conventional autoregressive policies trained with MSE loss, CARP employs a Cross-Entropy objective that preserves the potential for multi-modal predictions.
Compared to the Diffusion Policy~\cite{chi2023diffusionpolicy}'s ability to model multi-modality via DDPM's integration over stochastic differential equations~\cite{ho2020denoising}, CARP adopts a more direct yet effective alternative that achieves comparable multimodal expressiveness.
Nevertheless, the role of multi-modality in visuomotor policy learning remains underexplored. 
Many current benchmark tasks either do not require diverse output distributions or tend to induce overfitting to a single prediction path. 
Future research should investigate the necessity of multi-modal reasoning in robotic decision-making and further harness CARP’s capacity to model action diversity.

\textbf{Third}, CARP’s adoption of the GPT-style paradigm opens up promising yet unexplored possibilities. 
Beyond the flexibility already demonstrated, the contextual understanding capabilities inherent in GPT-style architectures~\cite{radford2019language} suggest that CARP could be extended to support multi-modal inputs~\cite{alayrac2022flamingo, radford2021learning} like tactile and auditory information 
and address robotic tasks requiring long-term dependency reasoning~\cite{ahn2022can}.
Moreover, its inherent capacity for in-context learning suggests strong potential for generalization under few-shot and zero-shot learning settings~\cite{brown2020language}, making it a compelling foundation for more adaptive and versatile visuomotor policies.

\textbf{Finally}, but not exhaustively, the scaling potential of CARP presents a promising avenue for future exploration.
The scaling laws established in existing GPT-style models~\cite{kaplan2020scaling} could be seamlessly applied to CARP, suggesting that increasing model capacity and leveraging larger pre-training datasets could lead to substantial performance gains. 
Furthermore, recent advances in Vision-Language-Action (VLA)~\cite{brohan2023rt, kim2024openvla, li2023vision} models present a promising opportunity to integrate CARP into such frameworks. Such integration could further demonstrate CARP’s scalability and its potential for general-purpose embodied intelligence.

\section{Coarse-to-Fine Inference}
\label{sec:suppl_infer_process}

Unlike the training process, the inference process predicts token maps of the action sequence across different scales in an autoregressive \textit{next-scale}, \textit{coarse-to-fine} manner without teacher forcing, as illustrated in \cref{fig:suppl_frame_infer}. Additionally, kv-caching is employed to eliminate redundant computations.

\begin{figure}[ht]
  \centering
   \includegraphics[width=1.0\linewidth]{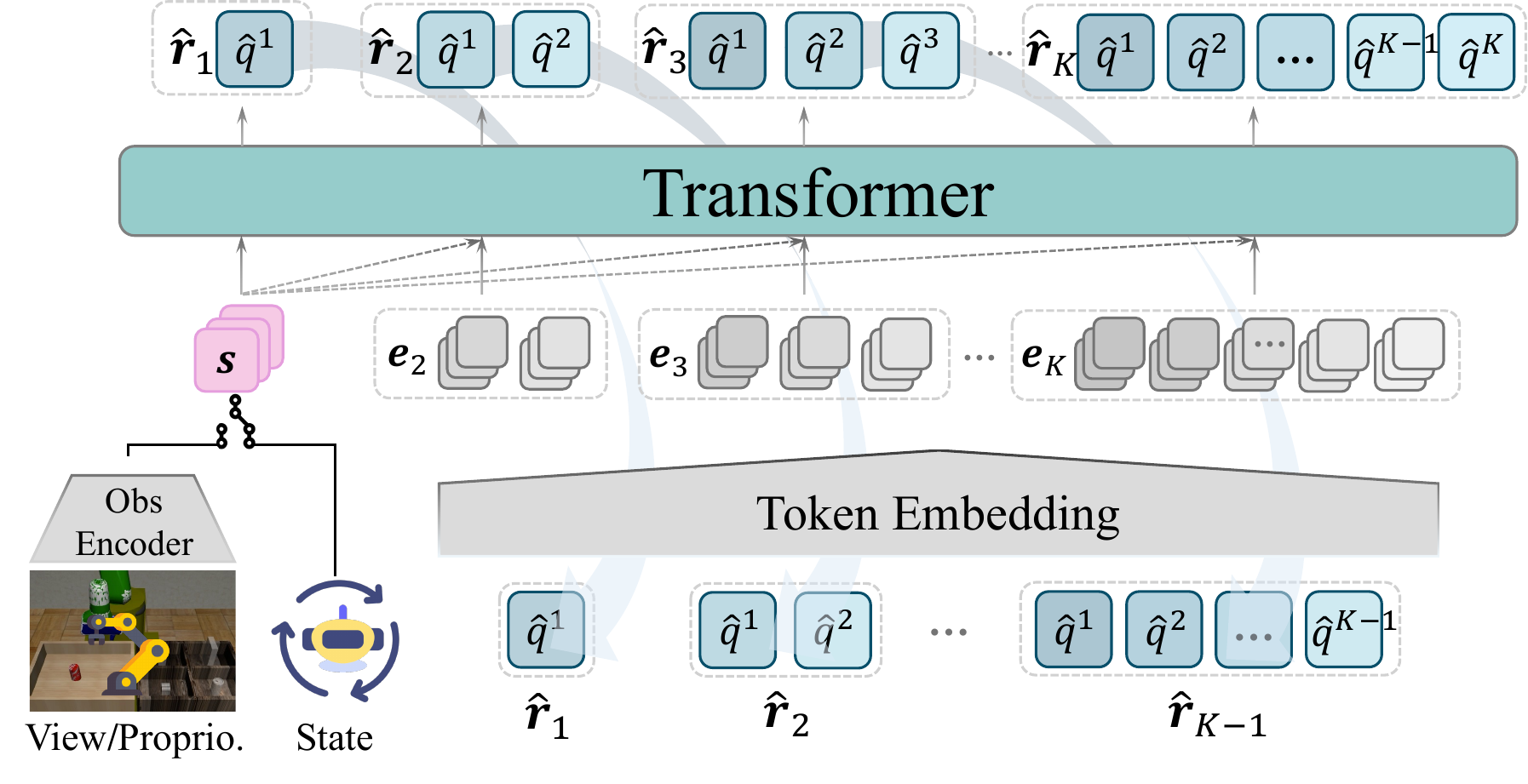}
   \caption{
   \textbf{Coarse-to-Fine Autoregressive Inference.}
    During inference, we leverage kv-caching to enable \textit{coarse-to-fine} prediction without the need for causal masks. 
    The full set of token maps, $\boldsymbol{r}_{1:K}$, is collectively decoded by the action \textit{multi-scale} VQVAE into executable actions for the robotic arm.
   }
   \label{fig:suppl_frame_infer}
\end{figure}

\section{Definition of Autoregressive Policy}
\label{sec:suppl_def_ar}

Autoregressive policies naturally capitalize on the efficiency and flexibility of autoregressive models. Initial works from a reinforcement learning perspective applied models like Transformers to predict the next action using states or rewards as inputs~\cite{chen2021decision,janner2021offline}, as shown in \cref{fig:suppl_intro_autoreg} and formalized as 
\begin{equation}
    p\left(\boldsymbol{a}_t,\boldsymbol{a}_{t+1},...,\boldsymbol{a}_{t+H-1}\right) = \prod^{t+H-1}_{k=t}p\left(\boldsymbol{a}_k|\boldsymbol{a}_{k-H:k-1},\boldsymbol{s}_{k-H:k}\right),
\label{eq:suppl_autoreg_pred}
\end{equation}
where $\boldsymbol{s}_{k-H:k-1}$ represents the states or observations corresponding to the previous actions $\boldsymbol{a}_{k-H:k-1}$, and $\boldsymbol{s}_{k}$ is the current state or observation.
Following this paradigm, several subsequent works~\cite{reed2022generalist,jiang2022vima,thomas2023plex} employ autoregressive models to predict one action at a time during inference.

\begin{figure}[ht]
  \centering
   \includegraphics[width=0.9\linewidth]{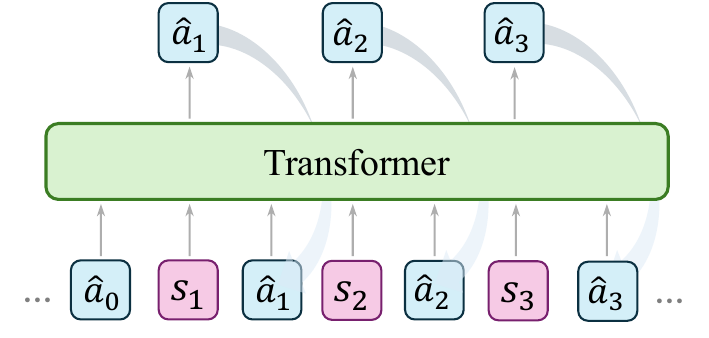}
   \caption{
   \textbf{Conventional Autoregressive Policy.}
    In reinforcement learning, conventional autoregressive policies generate action tokens sequentially, where each token is predicted based on the previously generated tokens. This differs from the action chunking prediction (see Sec. 2.2 of the main paper).
   }
   \label{fig:suppl_intro_autoreg}
\end{figure}

More recently, the concept of action chunking~\cite{lai2022action}, derived from neuroscience, has demonstrated notable benefits for imitation learning ~\cite{shafiullah2022behavior,cui2022play,zhao2023learning,lee2024behavior}. In action chunking, individual actions are grouped and executed as cohesive units, leading to improved efficiency in storage and execution, as depicted in
Fig. 2a of the main paper.
This paradigm extends the capabilities of GPT-style decoders by modifying them to generate chunks of actions in one forward pass, replacing the traditional single-step autoregressive operation with a multi-token, pseudo-autoregressive process. The action generation process for this paradigm is described as
\begin{equation}
    p\left(\boldsymbol{a}_t, \boldsymbol{a}_{t+1}, \dots, \boldsymbol{a}_{t+H-1}\right) = \prod^{t+H-1}_{k=t} p\left(\boldsymbol{a}_k | \boldsymbol{s}_O\right),
\label{eq:suppl_ar_pred}
\end{equation}
where $O$ is the historical horizon. The model predicts the entire action sequence in one forward pass without strictly adhering to step-by-step autoregressive operations.

Given the significant performance improvements enabled by action chunking, we adopt this multi-token, one-forward-pass framework throughout the article and experiments when referring to Autoregressive Policy (AP).

\section{Efficiency Concerns}
\label{sec:suppl_efficiency_concern}

Efficiency, as discussed throughout this paper, specifically refers to inference efficiency—the ability of CARP to generate actions significantly faster than diffusion-based policies during deployment. 
While efficiency can be examined from various perspectives, we focus on three key aspects relevant to CARP: inference efficiency, training efficiency, and data efficiency, each of which is analyzed in detail below.

\begin{table}[ht]
    \centering
    \begin{tabular}{l|cc|c}
    \multirow{2}[1]{*}{\textbf{Task}} & 
    \multicolumn{2}{c|}{\it{CARP}}& 
    \multirow{2}[1]{*}{\it{DP}} \\
        & Predict & Decode &   \\
    \hline
    Can  & 2.279 s  & 0.639 s & 34.79 s\\
    Square  & 2.597 s & 0.679 s & 35.62 s\\
    \end{tabular}
    \caption{
    \textbf{Inference Efficiency Comparison.} 
    We report the time consumption for the \textit{coarse-to-fine} prediction phase and the subsequent action decoding phase over 400 timesteps of action generation. The results indicate that in CARP, the majority of inference time is allocated to the prediction step, whereas the decoding process is completed within a short duration.
    }
    \label{tab:suppl_two_stage_inference_speed}
\end{table}

\textbf{Inference Efficiency.} 
We analyze the inference efficiency of CARP’s two-stage process. During inference, CARP first predicts action tokens in a \textit{coarse-to-fine} manner, followed by a single forward pass to decode all token maps into executable actions. 
Since token maps are collected during the prediction phase, and decoding requires only a single forward computation, the majority of the computational cost is incurred during prediction, while action decoding remains relatively lightweight. 
This is empirically validated by the results in \cref{tab:suppl_two_stage_inference_speed}. Compared to DP, CARP achieves significantly faster inference by eliminating the iterative denoising steps required by diffusion-based policies, instead directly predicting actions as a low-dimensional generation problem.

\textbf{Training Efficiency.} 
We analyze training efficiency through convergence behavior and time consumption. 

While training convergence depends on task complexity and hyperparameter configurations, both CARP and DP exhibit stable learning dynamics under our respective settings. 
To accommodate the architectural differences of CARP, we employ slightly different training configurations from those used for DP. 
As shown in \cref{fig:suppl_training_loss_curve}, under our experimental settings, both DP and CARP achieve good convergence within the same number of training epochs.
While convergence speed may differ due to structural and training differences, it does not inherently indicate superiority in model design. Instead, both CARP and DP demonstrate reliable training behavior under their respective settings.

\begin{figure}[ht]
  \centering
   \includegraphics[width=1.0\linewidth]{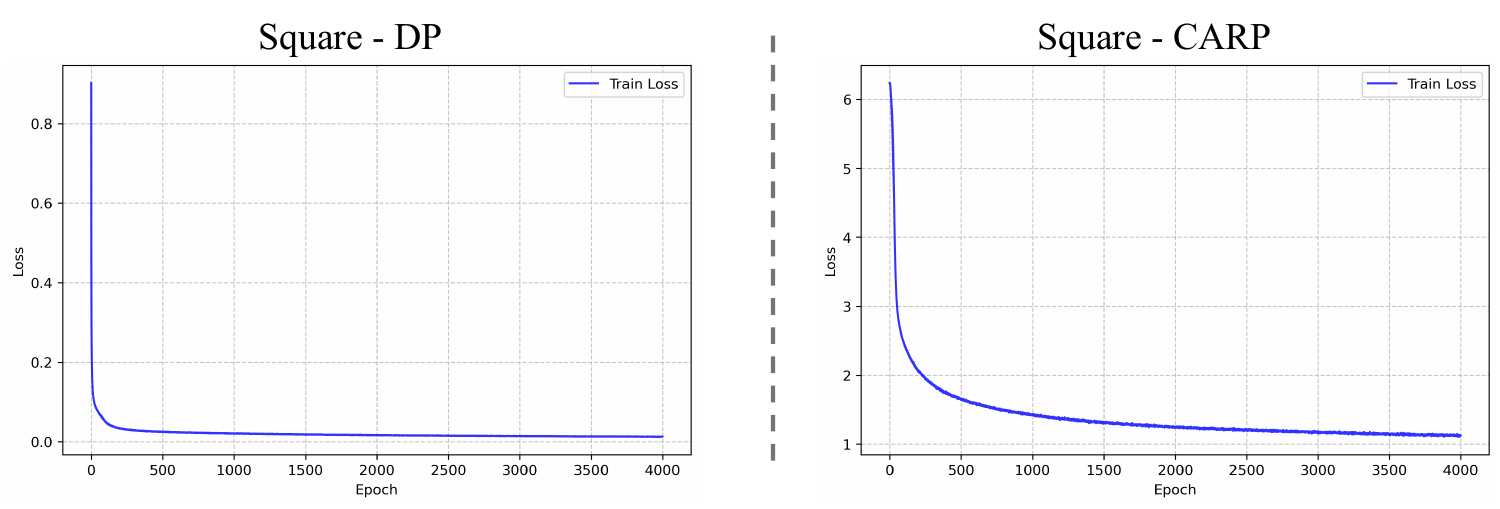}
   \caption{
    \textbf{Training Efficiency on Convergence Analysis.}
    With different training configurations, both DP and CARP converge effectively within 4000 epochs in the state-based Square task.
   }
   \label{fig:suppl_training_loss_curve}
\end{figure}

\begin{figure}[ht]
  \centering
   \includegraphics[width=1.0\linewidth]{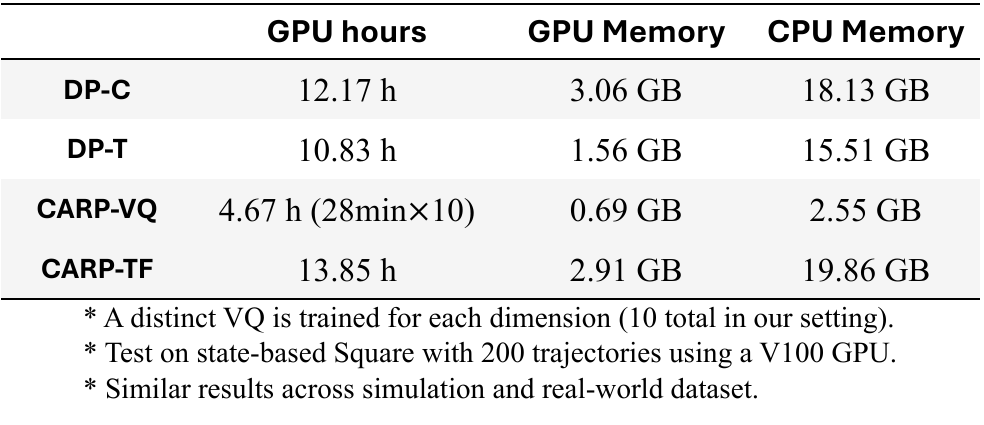}
   \caption{
    \textbf{Training Efficiency on Time Consumption.}    
    Comparison of training time on the Square task. For CARP, we separately report the tokenizer training time (with 10 VQ-VAEs, one per action dimension) and policy learning time.
   }
   \label{fig:suppl_training_efficiency}
\end{figure}

In terms of wall-clock training time, a comparison is provided in \cref{fig:suppl_training_efficiency}. 
The total training cost of CARP is comparable to that of DP when considering the policy learning stage (\texttt{CARP-TF}) alone. 
Although CARP introduces an additional tokenizer pretraining phase (\texttt{CARP-VQ})—where separate VQ-VAEs are trained for each action dimension—the cost is amortized across tasks and environments. 
In particular, when trained on sufficiently diverse data (e.g., multi-task settings in both simulation and the real world), the tokenizers become reusable, substantially reducing the overall training burden in practical deployments.

\textbf{Data Efficiency.}
We further assess the data efficiency of CARP by evaluating its performance under varying amounts of training data. Specifically, we investigate whether CARP can maintain strong performance when trained with limited trajectories, indicating robustness to data scarcity.
As shown in \cref{fig:suppl_data_efficiency}, we compare CARP with baseline policies (following implementations from~\cite{chi2023diffusionpolicy}) on the state-based Square task, using training datasets ranging from 200 to 30 trajectories. CARP consistently outperforms the baselines across all data regimes, demonstrating its superior data efficiency and reduced reliance on large-scale datasets.

\begin{figure}[htbp]
    \centering
    \includegraphics[width=0.82\linewidth]{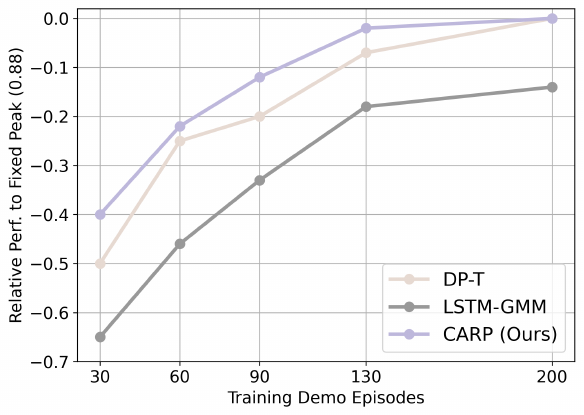}
    \caption{
    \textbf{Data Efficiency Analysis.}
    Performance comparison under varying training dataset sizes on the Square task. Each policy is trained following its official best-practice settings. CARP consistently outperforms the baselines at all data scales.
    }
    \label{fig:suppl_data_efficiency}
\end{figure}

\section{Comparative Analysis of CARP and AR.}
\label{sec:suppl_analysis_carp_w_ar}

CARP outperforms traditional autoregressive (AR) models in terms of success rate while maintaining high computational efficiency. 
CARP adopts a straightforward action tokenizer and leverages a GPT-style transformer for prediction, similar to standard AR policies. 
However, instead of conventional \textit{next-token} prediction, CARP introduces a paradigm shift towards \textit{next-scale} prediction. Despite these seemingly minor modifications, CARP achieves substantial performance gains. 
In this section, we analyze the key factors contributing to this improvement.

\textbf{Temporal Locality.} 
CARP encodes action sequences into a latent space in its first stage. Specifically, it employs 1D convolution to explicitly capture the local correlations within actions, facilitating a more effective learning of temporal dependencies—something that step-by-step action modeling struggles to achieve. 
As depicted in the magnified region of \cref{fig:suppl_vqvae_recon_mse}, encoding actions into a latent space enhances the smoothness of predictions while simultaneously denoising raw actions. 
This encoding process enables the model to capture similarities and overarching trends across contiguous actions, leveraging temporal locality to its advantage. 
While recent work on action chunking~\cite{zhao2023learning} has highlighted the significance of temporal locality, existing \textit{next-token} prediction models still suffer from weakened action dependencies due to their traditional independent action output mechanisms.

\textbf{Global Structure.} 
CARP represents action sequences across multiple scales and predicts actions in a \textit{coarse-to-fine} manner. 
The coarser scales compress the sequence using fewer tokens, promoting the learning of global action patterns. 
This hierarchical representation explicitly models the overall structure of action sequences—an aspect that traditional unidirectional \textit{next-token} prediction struggles to capture.
By progressively refining actions from high-level to low-level representations, CARP enhances sequence stability and mitigates the risk of producing erratic, inconsistent motions, leading to more precise execution.

\textbf{Action Scalability.} 
Similar to the approach used in Diffusion Policy, encoding action sequences improves the scalability of action generation. 
By encoding action sequences, CARP enables flexible adjustments to sequence length with minimal modifications to the model architecture, offering greater adaptability across different tasks and environments.

\begin{figure}[htbp]
    \centering
    \includegraphics[width=1.0\linewidth]{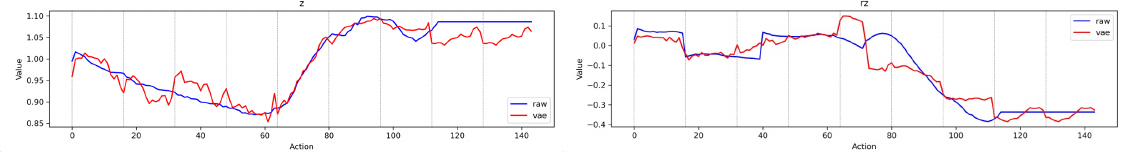}
    \caption{
    \textbf{Raw vs. Reconstructed Actions (Jointly Trained VQ-VAE).}
    Flattening and jointly encoding all action dimensions into a single VQ-VAE leads to poor reconstruction, as shown by the large discrepancy between raw (blue) and reconstructed (red) trajectories. This highlights the limitation of naive joint encoding.
    }
    \label{fig:vis_comb_vae}
\end{figure}

\begin{figure}[htbp]
    \centering
    \includegraphics[width=1.0\linewidth]{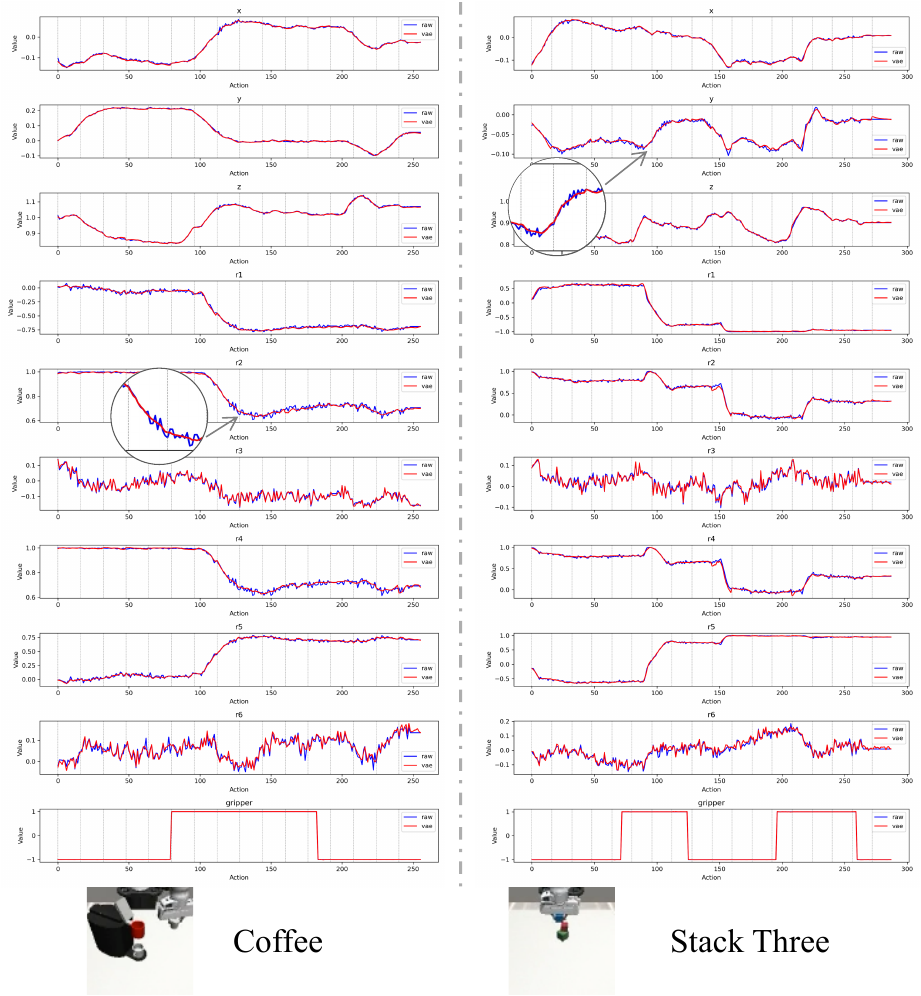}
    \caption{
    \textbf{Comparison of Raw and Reconstructed Actions.}     
    Comparison across 10 action dimensions in the Coffee and Stack Three tasks. Reconstructed actions (red) closely align with raw signals (blue), preserving structural patterns while smoothing the sequences and filtering out noise (see magnified region), highlighting the effectiveness of our \textit{multi-scale} tokenization.
    }
    \label{fig:suppl_vqvae_recon_mse}
\end{figure}

\textbf{Precision in Encoding.} 
Adapting the \textit{multi-scale} VQ-VAE to action space necessitates careful architectural design.
Rather than flattening the action trajectory into an image-like structure and training a single joint VQ-VAE—which leads to unstable and less interpretable tokens (\cref{fig:vis_comb_vae})—we instead encode each action dimension independently, using a dedicated VQ-VAE per dimension.
As illustrated in \cref{fig:suppl_vqvae_recon_mse}, the \textit{multi-scale} tokenization process ensures that generated action sequences closely match the raw inputs, exhibiting nearly identical trajectory lines while yielding smoother motions.
This demonstrates that our action tokenization approach effectively preserves the fidelity of original action sequences. 
Moreover, the enhanced success rates observed in the experiments presented in the main paper further validate the accuracy and effectiveness of CARP’s design.

\section{Additional Baselines Comparison}
\label{sec:suppl_add_baselines_comp}

In addition to the baseline policies discussed in the main paper, we further compare CARP with enhanced versions of each category: VQ-BET~\cite{lee2024behavior}, an improved variant of BeT~\cite{shafiullah2022behavior}, and Consistency Policy~\cite{prasad2024consistency}, which reduces sampling steps to improve inference efficiency. Both are implemented using their official codebases with recommended settings.
As shown in~\cref{tab:addi_baselines}, CARP achieves consistently higher success rates across all tasks, while maintaining competitive inference time. These results highlight CARP as a promising design that combines strong task performance with high inference efficiency.

\begin{table}[htbp]
\centering
\begin{minipage}[t]{0.68\linewidth}
    \centering
    \tablestyle{5pt}{1.0}
    \setlength\tabcolsep{4pt}
    \def\w{20pt}
    \scalebox{1}{
        \begin{tabular}{l|ccccc}
        Policy & p1 & p2 & p3 & p4 & Inf.T(s) \\
        \hline
        ConsisP & 0.99 & 0.96 & 0.95 & 0.93 & 2.31 \\
        VQ-BeT  & 0.96 & 0.92 & 0.87 & 0.71 & 1.48 \\
        \rowcolor[rgb]{ .949,  .949,  .949} 
        CARP    & \textbf{1.00} & \textbf{1.00} & \textbf{0.98} & \textbf{0.98} & 2.01\\
        \end{tabular}
    }
\end{minipage}
\hfill
\begin{minipage}[t]{0.28\linewidth}
    \centering
    \tablestyle{5pt}{1.0}
    \setlength\tabcolsep{4pt}
    \def\w{20pt}
    \scalebox{1}{
        \begin{tabular}{cc}
         Push-T & Inf.T(s) \\
        \hline
        0.80 & 2.93 \\
        0.72 & 1.70 \\
        \rowcolor[rgb]{ .949,  .949,  .949} 
        \textbf{0.88} & 2.66 \\
        \end{tabular}
    }
\end{minipage}
\caption{
\textbf{State-Based Kitchen and Push-T Results.}
We compare CARP with VQ-BET~\cite{lee2024behavior} and Consistency Policy~\cite{prasad2024consistency} under identical settings. CARP consistently outperforms both baselines in success rate, while offering competitive inference times.
}
\label{tab:addi_baselines}
\end{table}

\section{Failure Analysis}
\label{sec:suppl_failure_analysis}

In this section, we analyze common failure cases observed during experiments with both the diffusion-based policies and our proposed CARP.

\textbf{Accident Recovery.} A notable failure mode is the inability to recover from disturbances, as illustrated in~\cref{fig:suppl_failure_recovery_vis}. When tools are accidentally knocked over due to suboptimal action trajectories, the model struggles to generate appropriate recovery behaviors. This limitation arises because the policy is trained purely by imitating expert demonstrations, which do not account for such out-of-distribution failure scenarios. Addressing this issue requires further incorporation mechanisms for failure detection and recovery. 

\textbf{Hesitant Movements.}
Another common failure case involves the generation of jerky or oscillatory movements when the robot encounters two similar situations with only slight visual differences across consecutive timesteps, as shown in the first row of~\cref{fig:suppl_failure_hesitate_vis}.
This issue arises because the policy conditions its predictions on only the previous one or two observations, potentially overlooking long-term historical context. 
When faced with multiple plausible action choices under limited observations, the policy may produce ambiguous actions, leading to hesitation. 
Consequently, these hesitant movements can prevent the robot from meeting the success criteria, as shown in the bottom-right of~\cref{fig:suppl_failure_hesitate_vis}.

\begin{figure}[htbp]
    \centering
    \includegraphics[width=0.5\textwidth]{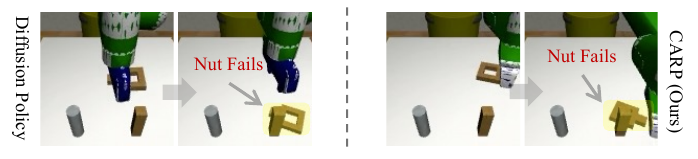}
    \caption{
    \textbf{Accident Recovery in the Square Task.} 
    When tools are accidentally knocked over, the robot struggles to recover due to its reliance on imitation learning from expert demonstrations, which lack exposure to such out-of-distribution failure cases.
    }
    \label{fig:suppl_failure_recovery_vis}
\end{figure}

\begin{figure}[htbp]
    \centering
    \includegraphics[width=0.35\textwidth]{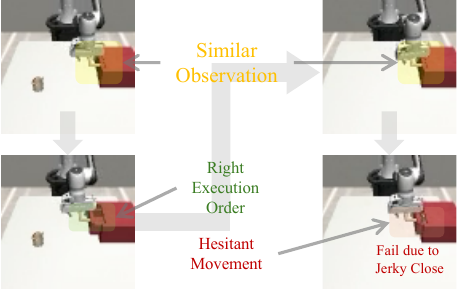}
    \caption{
    \textbf{Hesitant Movements in the Mug Task.}  
    During task execution, the robot may encounter visually similar observations at different timesteps. 
    Without leveraging long-term historical context, the policy may misinterpret these similarities and generate ambiguous actions, resulting in hesitant (jerky) movements. 
    In this failure case, after successfully closing the drawer, the policy perceives the scene as similar to the initial step and erroneously attempts to reopen it. This cycle of opening and closing continues indefinitely, leading to task failure.
    }
    \label{fig:suppl_failure_hesitate_vis}
\end{figure}

\section{Experiment Implementation Details}
\label{sec:suppl_implementation_details}

Here, we provide implementation details for the main experiments presented in the paper.

\textbf{Single-Task Simulation Experiment.}
For baseline models, we follow the same implementation and training configurations provided by Diffusion Policy.
For all state-based experiments, including the Kitchen and Push-T tasks, we uniformly set the observation horizon $O = 2$ and the prediction horizon $ H = 16 $ across all models.
For image-based experiments, we set $ O = 1 $ and $ H = 16 $ for better transferability to real-world scenarios. 
As per the benchmark, only the first 8 actions in the prediction horizon are executed, starting from the current step (see Suppl. I for further discussion).
For CARP, we first train an action VQVAE model
(see Sec. 3.1 of the main paper)
following~\cite{lee2022autoregressive}, using $V=512$, $C=8$, a batch size of 256, and 300 epochs per task.
Given a horizon $H = 16$, we design \textit{multi-scale} representations with scales of $1, 2, 3$ and $4$ to capture \textit{coarse-to-fine} information across the action sequence. 
We then train an autoregressive GPT-2 style, decoder-only transformer 
(see Sec. 3.2 of the main paper)
, based on~\cite{tian2024visual}, using the same training settings as the benchmark, with a batch size of 256 for state-based experiments (4000 epochs) and a batch size of 64 for image-based experiments (3000 epochs). 
We use Cross-Entropy loss during training, which preserves the model's sampling capability. During inference, we typically select the token with the highest probability at each scale. However, to visualize multi-modal behavior in the Push-T task, we sample the top-$k$ tokens at each scale (with $k{=}3$), allowing for diverse predictions with controlled randomness.

\textbf{Multi-Task Simulation Experiment.}
To enable multi-task generalization, CARP augments the single-task formulation by introducing a learnable 3-dimensional task embedding for each task. These embeddings, retrieved based on task indices, are concatenated with the observation sequence $\boldsymbol{s}$ and act as additional conditional inputs to the policy. In our experiments involving 8 tasks, this corresponds to an $8 \times 3$ embedding matrix.
We also use a moderately deeper decoder-only transformer in GPT-2 style. 
CARP is trained with a batch size of 512 for 200 epochs on an A100 GPU. Baseline models follow the same training settings as SDP~\cite{wang2024sparse}. This minimal modification enables CARP to adapt to multi-task learning seamlessly.

\textbf{Real-World Experiment.}
For both baselines and CARP, the input consists of current visual observations from the wrist and scene cameras (resolution: $120 \times 160$), as well as proprioceptive data from the robotic arm. We execute 8 predicted actions out of a horizon of 16 predictions. 
We train the diffusion policy for 3000 epochs with a batch size of 64. 
For CARP, we use the same visual policy structure as in the simulation tasks, training the \textit{multi-scale} action tokenizer for 300 epochs with a batch size of 256, and the \textit{coarse-to-fine} transformer for 3000 epochs with a batch size of 64.

\section{Consistent with Diffusion Policy}
\label{sec:suppl_imple_detail_consis_w_dp}

We adopt similar experimental settings with 1 or 2 observations, a prediction horizon of 16, and an executable action length of 8, following the standard setup used in Diffusion Policy (DP)~\cite{chi2023diffusionpolicy}.
It is important to note that our classical formulation introduces a minor discrepancy in the horizon definition compared to the implementation of DP.
Specifically, in DP's experimental setting, the horizon $H$ encompasses the past observed steps, meaning that the index of the current next predicted action is $O$, rather than 0. 
In contrast, as outlined in the formulation of Eq. (1) in the main paper, %
the horizon $H$ does not include past observations, with the first prediction step corresponding to the next time step. While the rationale behind this design remains unclear due to limitations in the author's understanding, we retain the horizon definition introduced by Diffusion Policy (DP)~\cite{chi2023diffusionpolicy} to ensure consistency in our experimental comparisons.

\section{Ablation Study on the Number of Scales}
\label{sec:suppl_ablation_k}

\begin{figure*}[htbp]
    \centering
    \includegraphics[width=0.75\linewidth, keepaspectratio]{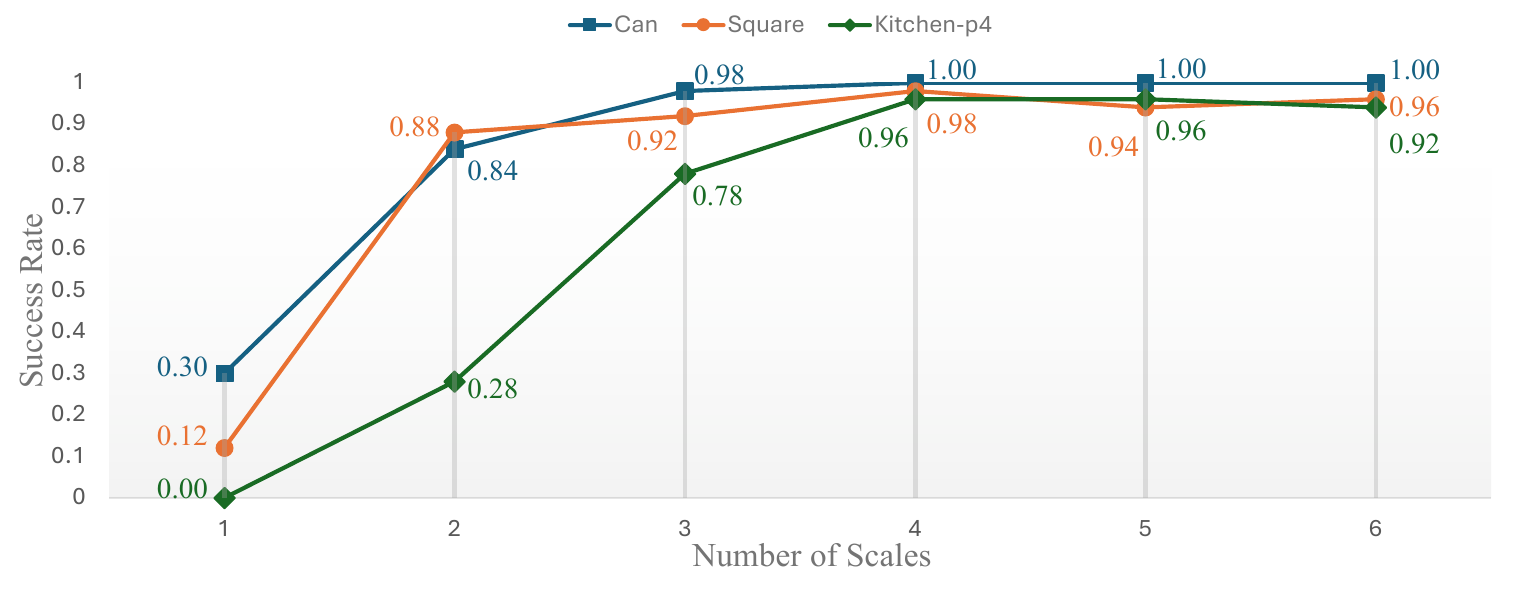}
    \caption{\textbf{Ablation Study on $K$.} We evaluate the performance of CARP across three tasks using six different scale configurations. Results indicate that when the number of scales exceeds 4, the model achieves optimal performance. Considering both model efficiency and performance, we set $K=4$ in all experiments throughout the paper.}
    \label{fig:suppl_ablation_scales}
\end{figure*}

To maintain consistency with Diffusion Policy, we set the action prediction horizon $H$ to 16 across all tasks. Given $H=16$, we adopt $K=4$ for all experiments. To further investigate the impact of $K$, we conduct an ablation study by varying $K$ from 1 to 6 on three representative tasks: Can, Square, and Kitchen (which requires executing four consecutive subtasks, thus we report success rate based on the final subtask, denoted as p4). All other experimental settings remain unchanged.

For each chosen number of scales $K$, the token map sizes at each scale level are defined as ${1, \dots, K}$. Notably, when $K=5$ and $K=6$, the scales slightly exceed the default feature map size. To ensure fair evaluation, we appropriately expand the feature map size to accommodate these settings.

As shown in \cref{fig:suppl_ablation_scales}, using fewer scales leads to insufficient action tokenization, resulting in less precise predictions. When $K=4$, the policy effectively meets task requirements. 
Further increasing $K$ results in stable performance with negligible fluctuations, indicating that the policy has likely reached its peak performance for the given tasks, with additional scaling providing little to no further benefit.

\section{Additional Real-World Experiment}
\label{sec:suppl_add_real_world}

We provide further visualization of the real-world experiments presented in the main paper. As shown in~\cref{fig:realworld_vis}, CARP generates smooth and successful trajectories for the \textit{Cup} and \textit{Bowl} tasks, with temporal progression illustrated from left to right.

\begin{figure}[ht]
  \centering
   \includegraphics[width=1.0\linewidth]{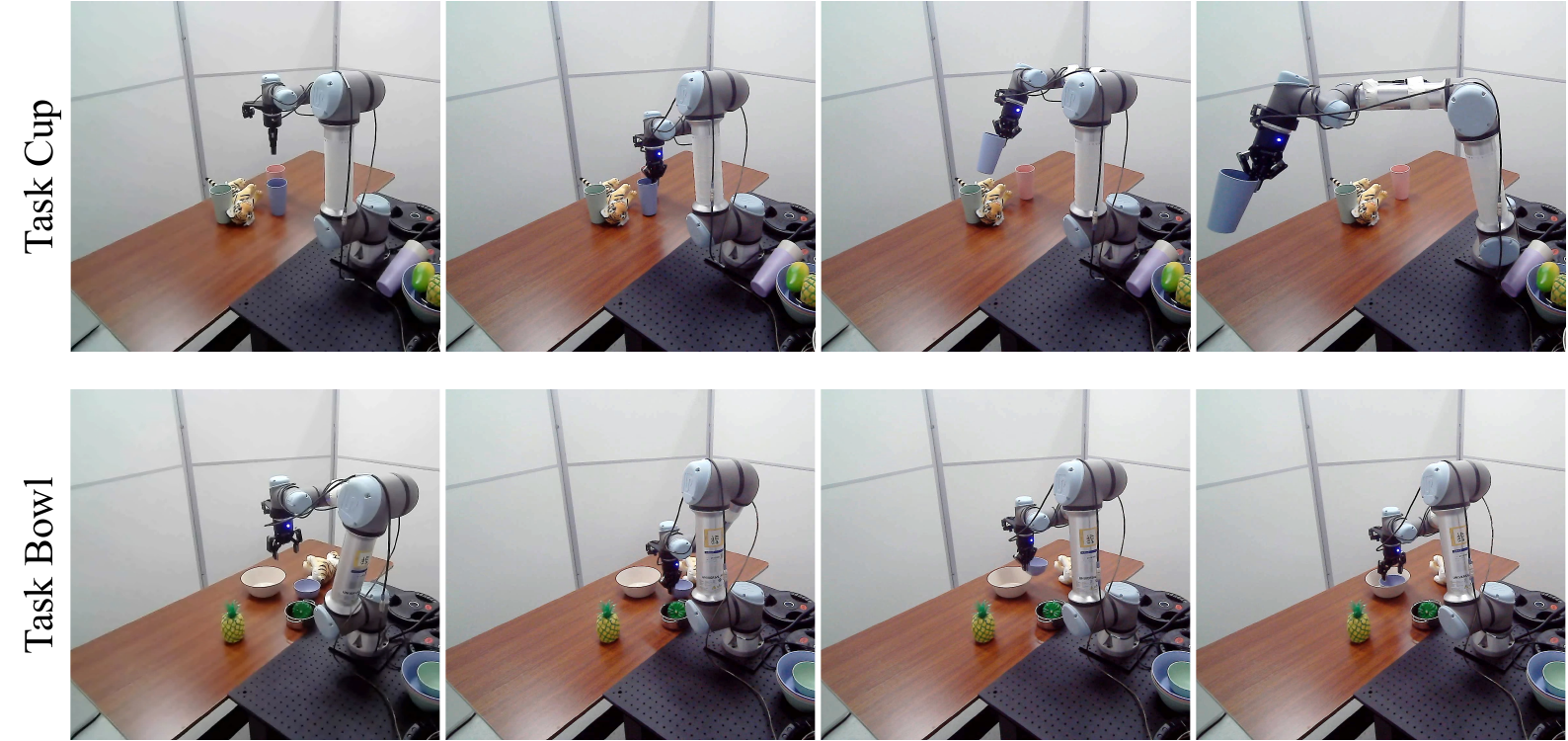}
   \caption{
   \textbf{Visualization of CAPR on Real-World Tasks.}
   CARP generates smooth and successful trajectories on the Cup and Bowl tasks, progressing from left to right.}
   \label{fig:realworld_vis}
\end{figure}

Beyond the real-world evaluations on a UR5e robot arm, we further deploy CARP on a distinct robotic embodiment: a 7-DoF Franka Emika Panda arm. For this, we adopt the challenging \textit{FurnitureBench} benchmark~\cite{heo2023furniturebench}, which comprises long-horizon, contact-rich manipulation tasks (e.g., pick, place, insert, screw, and flip), with episodes spanning up to 1000 steps (700 steps for \texttt{One\_Leg}). A corresponding standard simulator is also provided, as shown in~\cref{fig:suppl_furniture_vis}.

We first evaluate CARP and Diffusion Policy (DP) in simulation on three tasks, followed by real-world deployment of the \texttt{One\_Leg} task. 
All experiments are conducted in the state-based setting. 
To bridge the sim-to-real gap during real-world deployment, 6-DoF object poses are estimated using AprilTags provided by FurnitureBench~\cite{heo2023furniturebench}, enabling consistent state-based policy execution.

For simulation, we use 200 trajectories per task from~\cite{ankile2024imitation}. DP is trained using its official implementation~\cite{chi2023diffusionpolicy} with 100 DDPM denoising steps. 
CARP follows the same state-based, single-task setting. 
We evaluate success rates using 1024 rollouts per task for statistical stability. 
As shown in~\cref{tab:furniture_sim_results}, CARP achieves competitive performance while offering significantly lower inference time and fewer parameters.

\begin{figure}[htbp]
    \centering
    \includegraphics[width=1.0\linewidth]{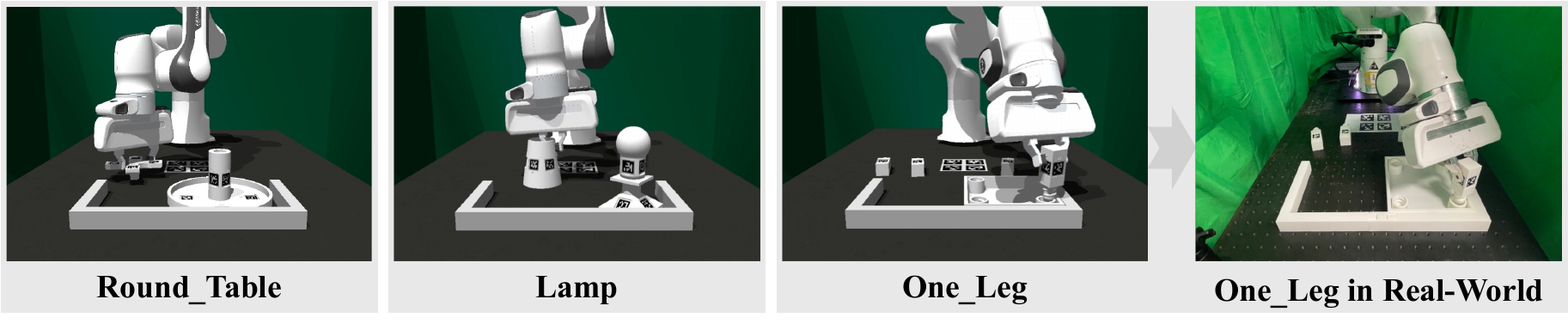}
    \caption{
    \textbf{FurnitureBench Tasks for Evaluation.}
    All three tasks are first evaluated in simulation. A corresponding real-world environment is then constructed to assess the performance in real-world.
    }
    \label{fig:suppl_furniture_vis}
\end{figure}

\begin{table}[htbp]
    \tablestyle{5pt}{1.0}
    \setlength\tabcolsep{4pt}
    \def\w{20pt}
    \scalebox{1}{
        \begin{tabular}{l|ccccc}
        Policy & One\_Leg  & Round\_Table & Lamp &  Inf.Time $\downarrow$ &  Params $\downarrow$ \\
        \hline
        DP-C & 39.62\%  & 5.76\% & 3.91\% & 74.05s & 66.06M \\
        \rowcolor[rgb]{ .949,  .949,  .949} 
        CARP & \textbf{43.75}\% & \textbf{6.25}\% & \textbf{4.30}\% & \textbf{6.29}s & \textbf{2.54}M \\
        \end{tabular}
    }
    \caption{
    \textbf{Simulation Results on FurnitureBench (State-Based).}
    All policies are trained under identical single-task settings. Compared to DP-C~\cite{chi2023diffusionpolicy} (with 100 DDPM denoising steps), CARP achieves comparable success rates while offering significantly lower inference time and parameter count.
    }
    \label{tab:furniture_sim_results}
\end{table}

We further evaluate CARP on the real-world \texttt{One\_Leg} task using 40 expert demonstrations collected via a 3D SpaceMouse (left panel of~\cref{fig:furniture_real_exp}).
The task involves complex rotations and contact-rich interactions, as illustrated in the right panel of~\cref{fig:furniture_real_exp}.
CARP achieves higher success rates across key stages, especially in precision-critical steps such as \textit{Insert}, demonstrating robust real-world performance.

\section{Analysis on Fine-Grained Manipulation}
\label{sec:suppl_fine_grained_manip}

Beyond standard pick-and-place tasks, which are relatively straightforward for robotic manipulation, tasks requiring fine-grained skills have garnered increasing attention. 
For example, Nut-Assembly and Threading from the MimicGen~\cite{mandlekar2023mimicgen} benchmark demand precise action generation to ensure successful task completion, as illustrated in \cref{fig:suppl_fine_grained_vis}. 
In Nut-Assembly, the robot must place a nut onto a designated peg, which is slightly larger in size. 
Furthermore, the Threading task requires the robot to insert a needle into a small hole on a tripod, a significantly more challenging task due to the minimal margin for error.
To evaluate CARP’s fine-grained manipulation capability, we compare it with the state-of-the-art Sparse Diffusion Policy (SDP)~\cite{wang2024sparse} in the multi-task setting, and Diffusion Policy (DP)~\cite{chi2023diffusionpolicy} in the single-task setting.
We evaluate success rates alongside the mean and variance of the distance between the ideal insertion centers of the fixed structures (peg, tripod) and the centers of the tools (nut, needle) at the moment of first contact.
A lower mean distance indicates higher action precision, while a smaller variance reflects the model's ability to consistently achieve accurate and stable manipulations.

As summarized in~\cref{tab:suppl_fine_grained_table} and~\cref{tab:suppl_fine_grained_table_single}, our \textit{coarse-to-fine} autoregressive prediction framework demonstrates strong performance in fine-grained tasks, achieving competitive results comparable to diffusion-based policies. 
Notably, CARP consistently achieves lower mean error and reduced variance across most tasks, regardless of whether in single-task or multi-task settings.
Moreover, CARP achieves these results with an inference speed that is \textbf{10} times faster than current diffusion-based policies, highlighting its efficiency and effectiveness in fine-grained robotic manipulation.

\section{Task Visualizations}
\label{sec:suppl_vis_tasks}
In this section, we provide visualizations of the tasks used in our experiments.  
For the single-task experiment, the corresponding visualizations are presented in \cref{fig:suppl_singletask_vis}.  
For the multi-task experiment, visualizations are shown in \cref{fig:suppl_multitask_vis}.  
For the long-horizon, multi-stage Kitchen experiment, we provide visualizations in \cref{fig:suppl_kitchentask_vis}, along with the sequential execution process in \cref{fig:suppl_kitchn_subtasks}.  
Finally, for real-world experiments, visualizations are included in \cref{fig:suppl_realworld_vis}.

\begin{figure*}[htbp]
    \centering
    \includegraphics[width=0.80\linewidth]{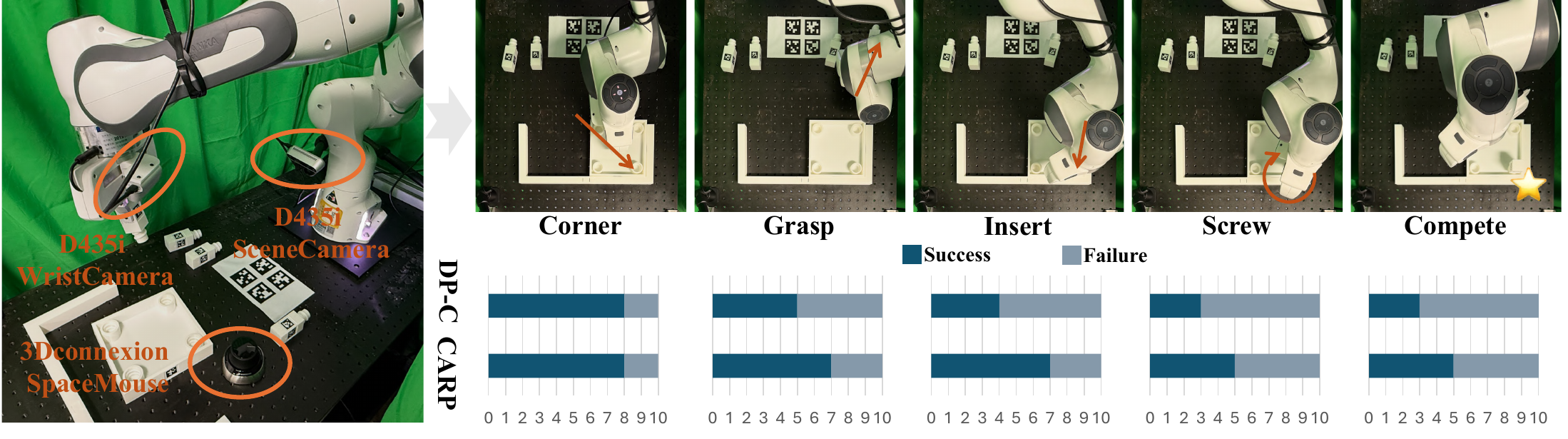}
    \caption{
    \textbf{Real-World Evaluation on the \texttt{One\_Leg} Task.}
    The left panel shows the real-world setup, while the right panel illustrates the execution process and stage-wise results (left to right). 
    Success rates at key stages are reported below. CARP produces smoother motions and outperforms baselines in precision-critical phases such as \textit{Grasp} and \textit{Insert}.
    }
    \label{fig:furniture_real_exp}
\end{figure*}

\begin{figure*}[htbp]
    \centering
    \includegraphics[width=0.70\linewidth]{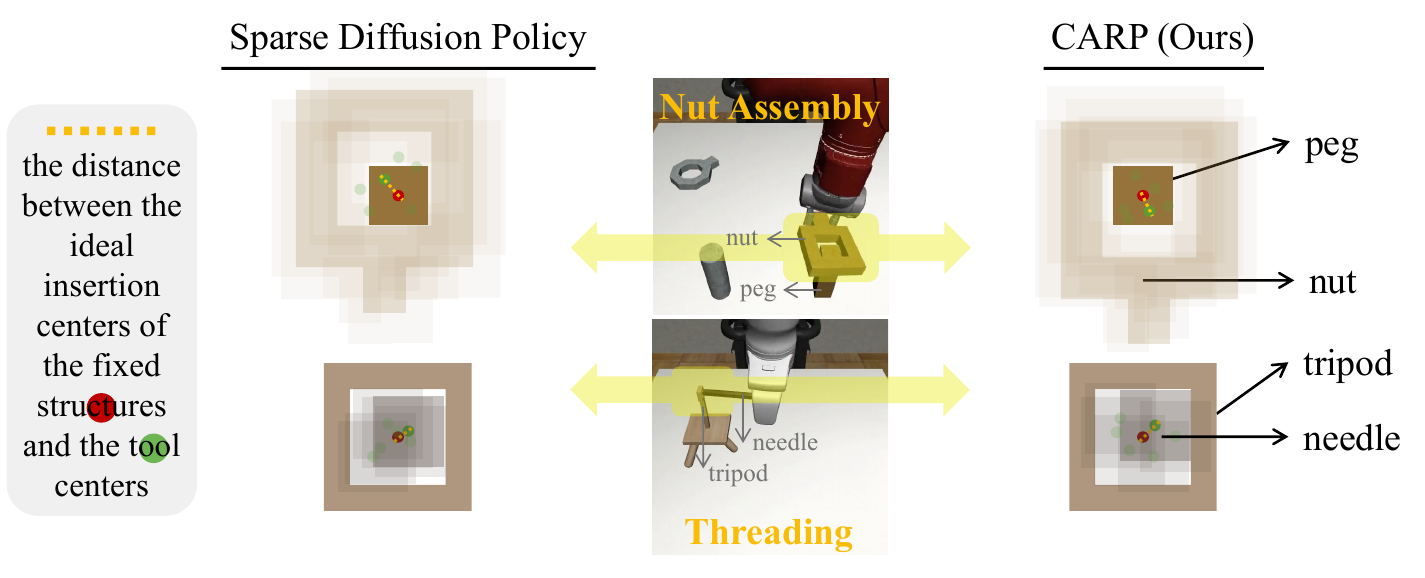}
    \caption{\textbf{Visualization of Fine-Grained Manipulation.} 
    We evaluate the precision of generated actions by measuring the distance between the ideal and actual contact centers, represented by the dotted yellow line. Experiments are conducted on Nut-Assembly (top row) and Threading (bottom row). The visualization highlights that CARP achieves a comparable level of precision to diffusion-based policies.
    }
    \label{fig:suppl_fine_grained_vis}
\end{figure*}

\begin{table*}[htbp]
    \centering
    \begin{tabular}{@{}l|c|ccc|ccc@{}}
    \multirow{2}[1]{*}{\textbf{Policy}} & 
    \multirow{2}[1]{*}{Inference Speed $\uparrow$} & 
    \multicolumn{3}{c|}{\it{Nut Assembly}} & 
    \multicolumn{3}{c}{\it{Threading} } \\
        &  & Success $\uparrow$ & Mean $\downarrow$ & Variance $\downarrow$  & Success $\uparrow$ & Mean $\downarrow$ & Variance $\downarrow$ \\
    \hline
    SDP & 8.20 hz & 0.54  & 7.70  & 2.36  & \textbf{0.70}  & \textbf{5.20}  & \textbf{1.25} \\
    \rowcolor[rgb]{ .949,  .949,  .949} 
    CARP & \textbf{118.50} hz & \textbf{0.66}  & \textbf{7.30}  & \textbf{1.68}  & \textbf{0.70}  & 5.50  & 1.36  \\
    \end{tabular}
    \caption{
    \textbf{Fine-Grained Manipulation Study on Multi-Task Setting.} 
    CARP demonstrates a high level of precision comparable to diffusion-based policies, as indicated by the similar mean and variance values.
    Additionally, CARP achieves a significant speed advantage, running over 10 times faster than diffusion-based approaches. This highlights CARP as a superior balance between performance and efficiency.
    }
    \label{tab:suppl_fine_grained_table}
\end{table*}

\begin{table*}[htbp]
    \centering
    \begin{tabular}{l|c|ccc|ccc}
    \multirow{2}[1]{*}{\textbf{Policy}} & 
    \multirow{2}[1]{*}{Inference Speed $\uparrow$} &
    \multicolumn{3}{c|}{\it{Nut Assembly}}& 
    \multicolumn{3}{c}{\it{Threading}}\\
    &   & Success $\uparrow$ & Mean $\downarrow$ & Variance $\downarrow$ & Success $\uparrow$ & Mean $\downarrow$ & Variance $\downarrow$ \\
    \hline
    DP-C & 10.13 hz & 0.80  & 5.20  & 1.86  & \textbf{0.88}  & 4.08  & 1.07  \\
    \rowcolor[rgb]{ .949,  .949,  .949} 
    CARP & \textbf{119.05} hz & \textbf{0.82}  & \textbf{5.12}  & \textbf{1.28}  & \textbf{0.88}  & \textbf{3.92}  & \textbf{0.94}  \\
    \end{tabular}
    \caption{\textbf{Fine-Grained Manipulation Study on Single-Task Setting.}
    To eliminate potential underfitting caused by multi-task training, we evaluate fine-grained tasks under single-task settings. 
    CARP achieves success rates on par with DP-C, while offering lower variance and over 10× faster inference. 
    These results highlight CARP’s ability to maintain high precision and stability with greater inference efficiency.
    }
\label{tab:suppl_fine_grained_table_single}
\end{table*}

\begin{figure*}[ht]
    \centering
    \includegraphics[width=0.95\linewidth, keepaspectratio]{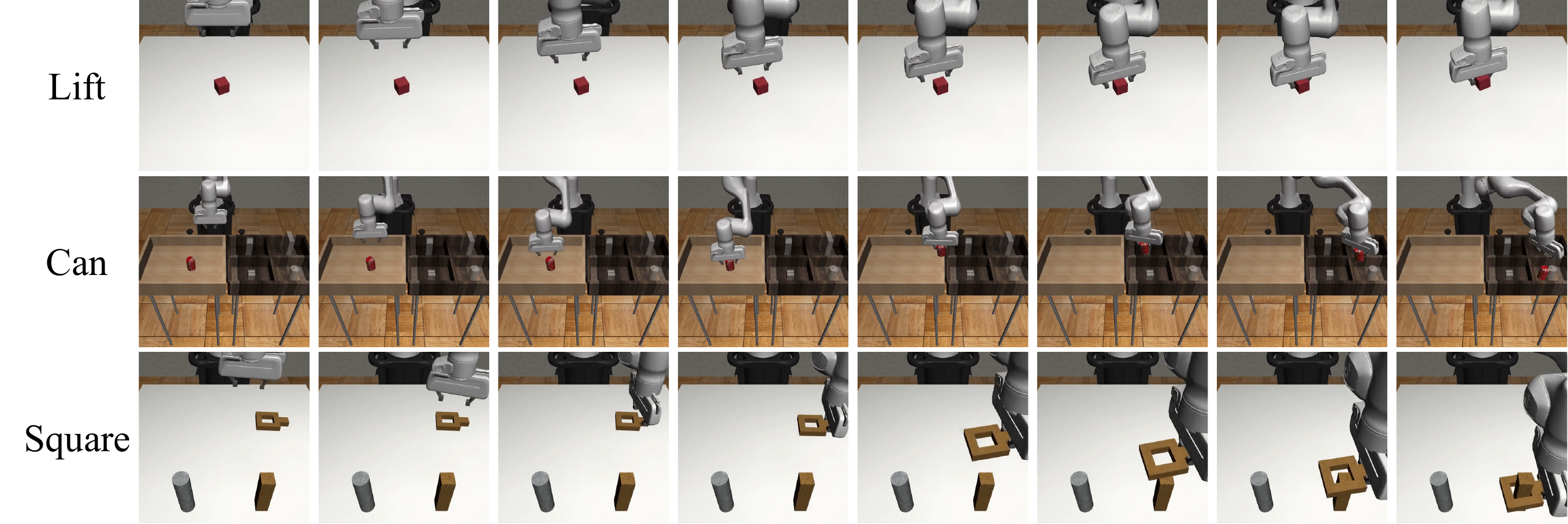}
    \caption{
    \textbf{Visualization of Tasks in Single-Task Experiment.}
    }
    \label{fig:suppl_singletask_vis}
\end{figure*}

\begin{figure*}[ht]
    \centering
    \includegraphics[width=0.95\linewidth, keepaspectratio]{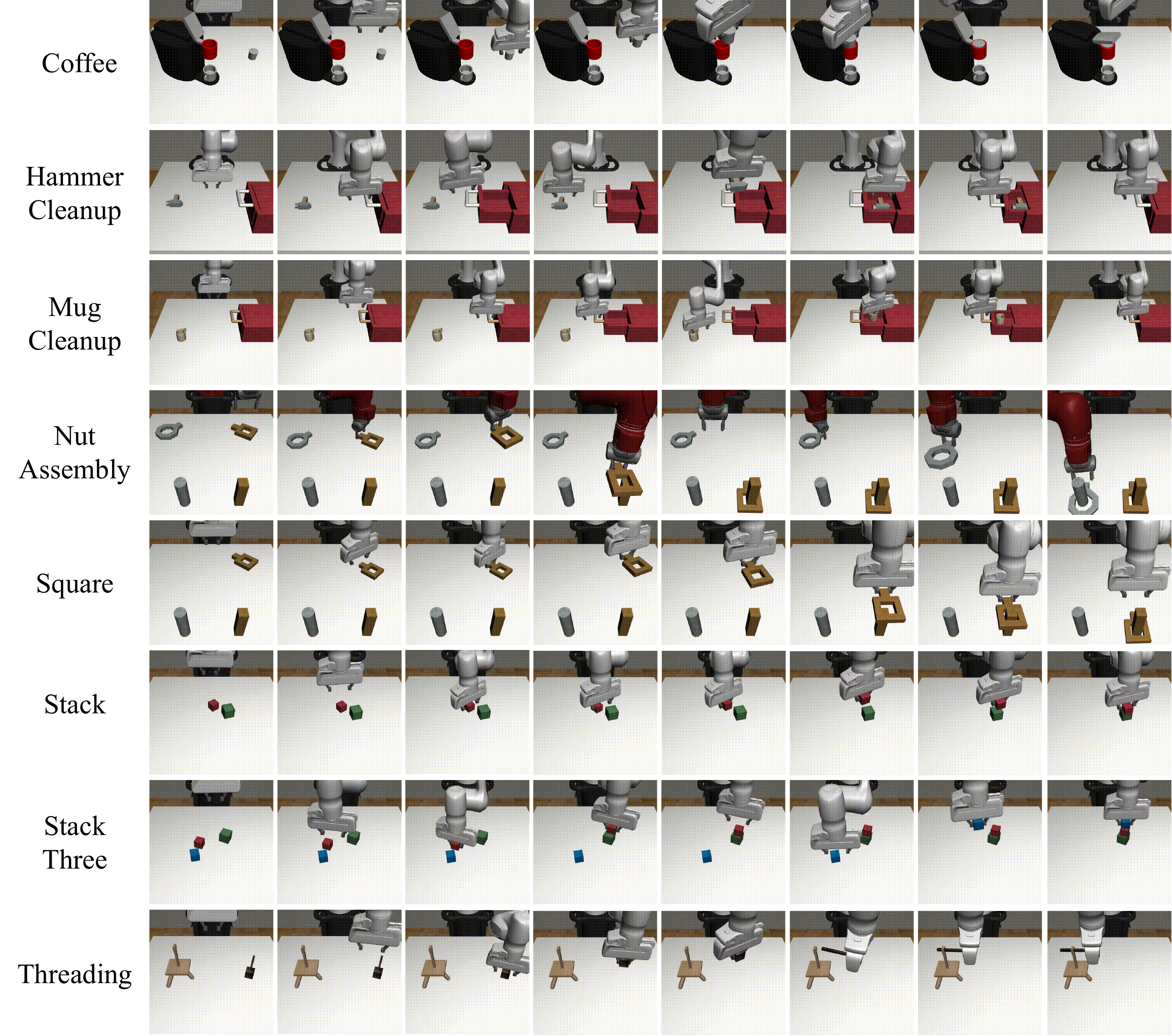}
    \caption{
    \textbf{Visualization of Tasks in Multi-Task Experiment.}
    }
    \label{fig:suppl_multitask_vis}
\end{figure*}

\begin{figure*}[ht]
    \centering
    \includegraphics[width=0.92\linewidth, keepaspectratio]{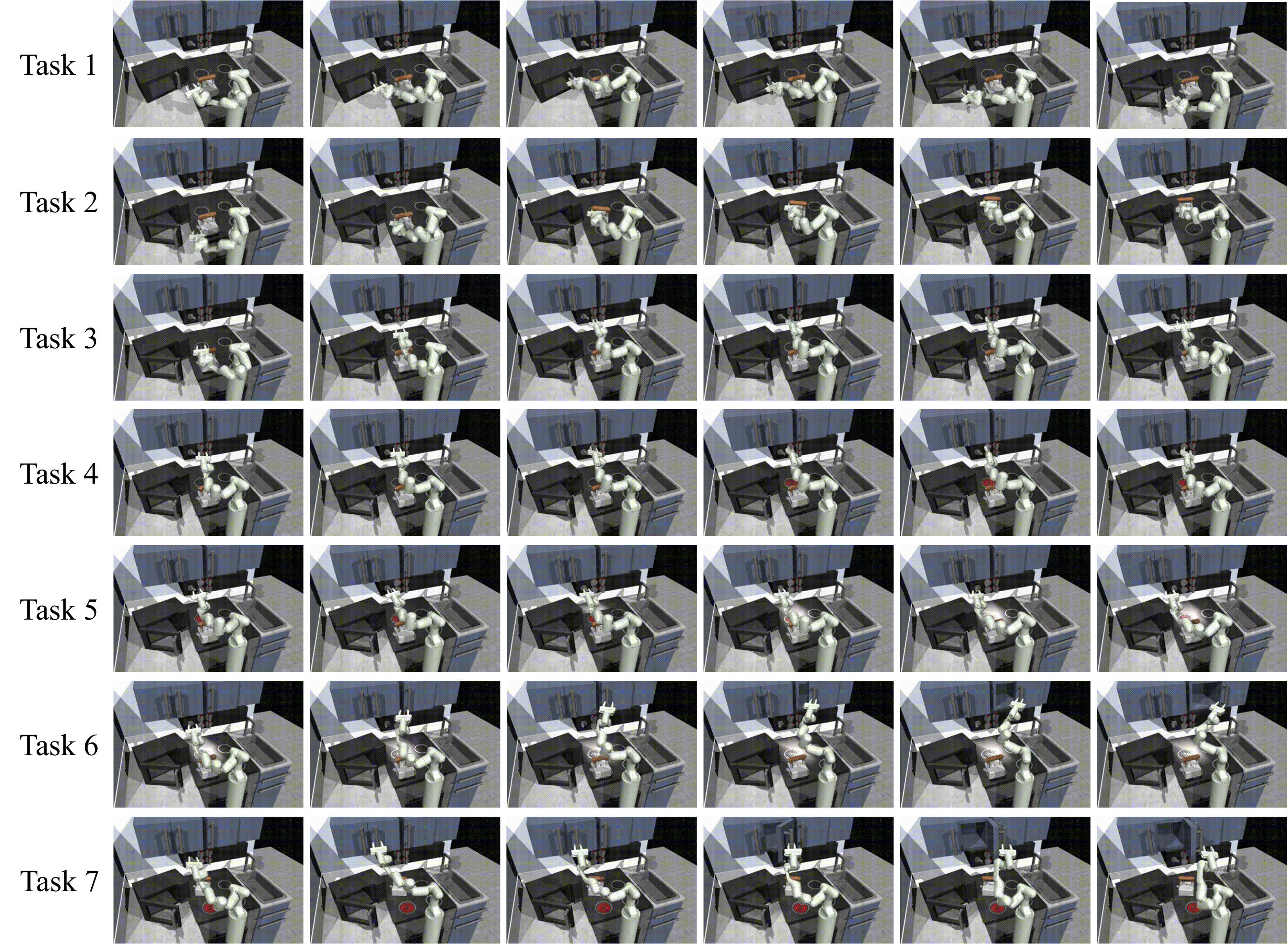}
    \caption{
    \textbf{Visualization of All Interaction Tasks in Kitchen Experiment.}
    }
    \label{fig:suppl_kitchentask_vis}
\end{figure*}

\begin{figure*}[ht]
    \centering
    \includegraphics[width=0.80\linewidth]{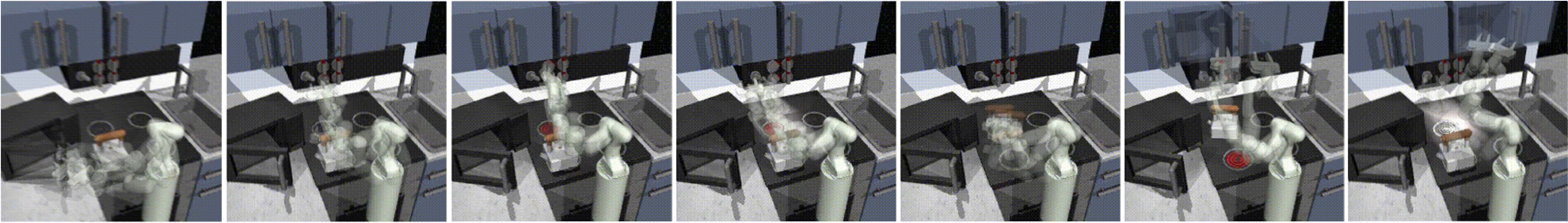} 
    \caption{
    \textbf{Visualization of the Consecutive Execution in Kitchen Experiment.}
    }
    \label{fig:suppl_kitchn_subtasks}
\end{figure*}

\begin{figure*}[ht]
    \centering
    \includegraphics[width=0.95\linewidth, keepaspectratio]{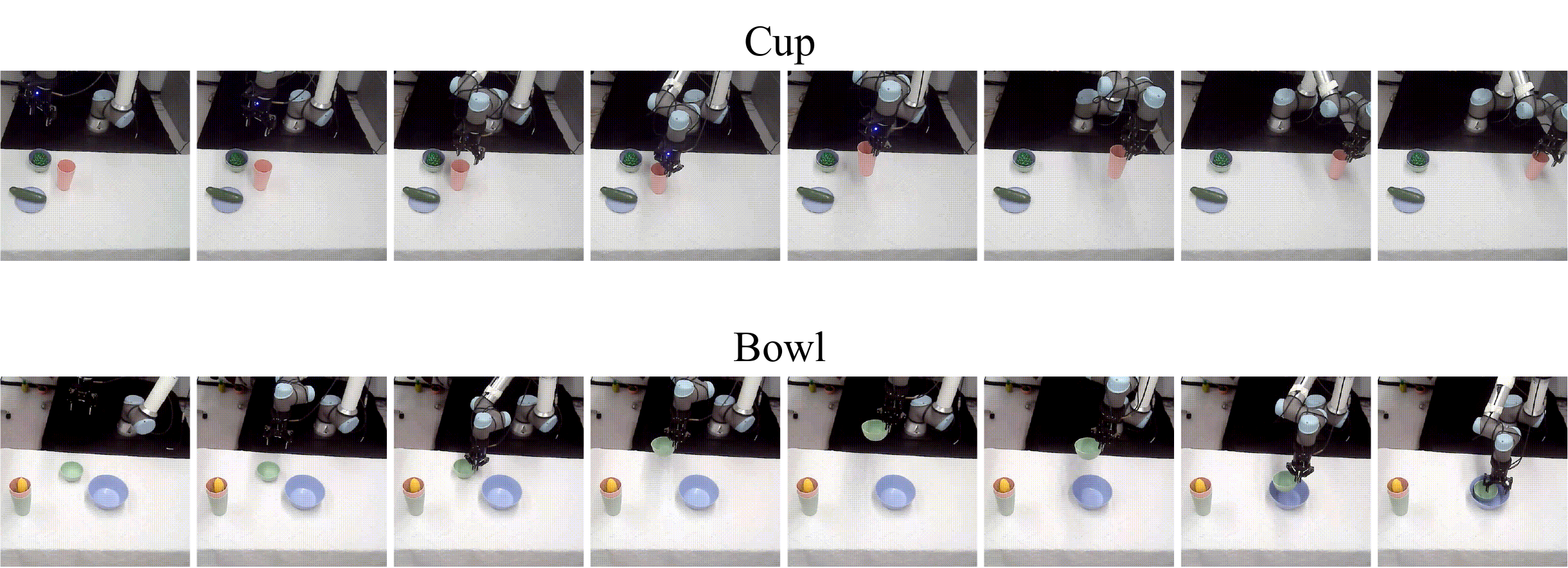}
    \caption{
    \textbf{Visualization of Tasks in Real-World Setup.}
    }
    \label{fig:suppl_realworld_vis}
\end{figure*}

 \fi

\end{document}